\documentclass{article}

\PassOptionsToPackage{numbers, compress}{natbib}
\usepackage[eandd, final]{neurips_2026}

\usepackage[utf8]{inputenc}
\usepackage[T1]{fontenc}
\usepackage[table]{xcolor}
\usepackage{hyperref}
\usepackage{url}
\usepackage{booktabs}
\usepackage{amsmath}
\usepackage{amssymb}
\usepackage{amsfonts}
\usepackage{graphicx}
\usepackage{multirow}
\usepackage{subcaption}
\usepackage{algorithm}
\usepackage{algpseudocode}
\usepackage{float}
\usepackage{makecell}
\usepackage{tabularx}
\usepackage{pifont}
\usepackage{microtype}
\usepackage{nicefrac}
\usepackage{caption}
\usepackage{placeins}
\usepackage{tcolorbox}
\tcbuselibrary{breakable}
\newcounter{promptbox}[section]
\renewcommand{\thepromptbox}{\thesection\arabic{promptbox}}
\newcommand{\cmark}{\ding{51}}
\newcommand{\xmark}{\ding{55}}
\definecolor{heavenlygold}{HTML}{E7D99F}

\usepackage[normalem]{ulem}
\newif\ifmarkup
\markupfalse  %
\ifmarkup
  \newcommand{\revadd}[1]{{\color{blue!80!black}#1}}
  \newcommand{\revdel}[1]{{\color{red!75!black}\let\revcite\cite\renewcommand{\cite}[1]{\mbox{\revcite{##1}}}\sout{#1}}}
  \newcommand{\revnote}[1]{{\color{red!75!black}\footnotesize\textit{[Removed: #1]}}}
\else
  \newcommand{\revadd}[1]{#1}
  \newcommand{\revdel}[1]{}
  \newcommand{\revnote}[1]{}
\fi
\newcommand{\revrep}[2]{\revdel{#1}\revadd{#2}}

\title{AmaraSpatial-10K: A Spatially and Semantically Aligned 3D Dataset for Spatial Computing and Embodied AI}
\author{%
  Mohammad Sadegh Salehi \\
  Zero One Creative \\
  \texttt{sadegh@01c.ai}
  \And
  \revrep{Alex Perkins}{Alexander J.P. Perkins} \\
  Zero One Creative \\
  \texttt{alex@01c.ai}
  \And
  \revrep{Igor Maurell}{Igor Pardo Maurell} \\
  Zero One Creative \\
  \texttt{igor@01c.ai}
  \And
  Ashkan Dabbagh \\
  Zero One Creative \\
  \texttt{ash@01c.ai}
  \And
  Raymond Wong \\
  Zero One Creative \\
  \texttt{raymond@01c.ai}
}

\begin{document}

\maketitle

\begin{abstract}
Web-scale 3D asset collections are abundant but rarely ready to ingest without cleanup,
suffering from arbitrary metric scaling, incorrect pivots, brittle geometry,
and incomplete textures, defects that limit their use in embodied AI,
robotics, and spatial computing. We present \textbf{AmaraSpatial-10K}, a
dataset of 10{,}071 synthetic 3D assets standardised for direct ingestion. Each asset ships as a metric-scaled, deterministically anchored
\texttt{.glb} with separated PBR maps, a convex collision hull, a paired
reference image, and multi-sentence text metadata. Alongside the dataset we
introduce a reusable evaluation suite for 3D asset banks, a continuous
Scale Plausibility Score (SPS), a keyword-based Concept Density metric, anchor-error
auditing, and a cross-modal CLIP coherence protocol, and apply it to
AmaraSpatial-10K alongside matched subsets of Objaverse, HSSD, ABO, and GSO.
We check the annotations independently of the LLM/VLM pipeline that produced them: measured ABO object heights fall inside our scale intervals 88.7--90.7\% of the time, and three non-author raters auditing a stratified sample of 314 assets judge 95.5\% plausibly scaled and 99.7\% correctly anchored.
AmaraSpatial-10K improves CLIP Recall@5 by $3.4\times$ over Objaverse
($0.612$ vs.\ $0.181$, median rank $267 \rightarrow 3$), achieves a $99.1\%$
settling-stability rate under Habitat-Sim in static placement and collision tests, with $\sim$$20\times$ lower per-load wall time, and as a drop-in Holodeck asset bank improves planned-scale matching and is preferred by blinded human raters in 27 of 40 paired scenes. Controlled ablations on the same asset bank attribute the
retrieval gain to description richness.
\end{abstract}
\section{Introduction}
\label{sec:intro}

\begin{figure}[ht]
    \begin{minipage}{0.55\linewidth}
        \centering
        \includegraphics[width=\linewidth]{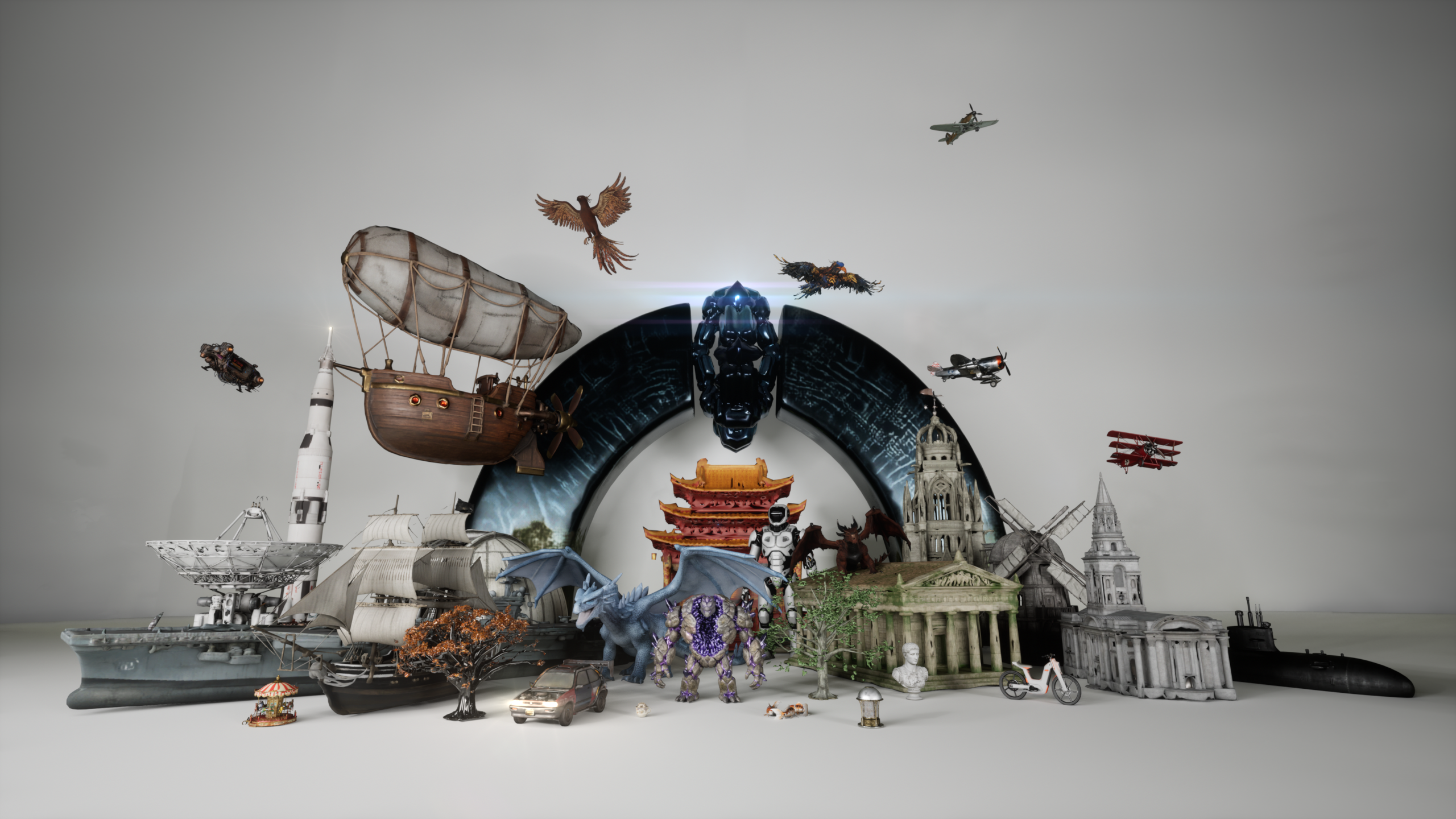}
    \end{minipage}\hfill
    \begin{minipage}{0.40\linewidth}
        \caption{\textbf{Representative assets from AmaraSpatial-10K.} Indoor
        objects, vehicles, architecture, creatures, and props, all released
        with metric scale, semantically correct anchoring, and PBR-ready
        materials under a shared spatial convention.}
        \label{fig:hero}
    \end{minipage}
\end{figure}

Recent 3D generative models~\cite{triposr, instantmesh, lrm, crm} synthesize
visually convincing meshes from a single image, but their outputs are rarely
ready for use as production assets. A generated chair may be 40~m tall, face
sideways relative to its canonical front, or place its pivot at the centroid
rather than the floor contact point. These are properties of how the models are built. LRM~\cite{lrm} and TripoSR~\cite{triposr} reconstruct in a normalised frame, rescaling outputs into a unit cube centred on the bounding box, so raw reconstructions carry no metric scale and are pivoted at the centroid. Image-conditioned generators also return arbitrary rotations instead of a canonical pose, which recent work treats as a research problem in its own right~\cite{orientationmatters,canoverse}. For embodied AI, robotics, and AR/VR
pipelines, these defects break placement, collision handling, physics, and
retrieval.

Existing 3D repositories trade scale, fidelity, or coverage against the
properties required for direct ingestion (\S\ref{sec:related}):
ShapeNet~\cite{shapenet} lacks materials and metric scale;
Objaverse~\cite{objaverse,objaversexl} offers volume but inconsistent quality;
GSO~\cite{gso} is high-fidelity but small. Filtering cannot recover properties
that were never authored. Metric scale, canonical orientation, and semantic
anchors are not inferable post-hoc from geometry alone.

This matters for two consumers. \textbf{Scene composition and
simulation} systems~\cite{habitat,igibson,procthor,holodeck,layoutgpt} consume
3D meshes at scale and require them to be metric, physically stable, and
semantically searchable. \textbf{Single-image-to-3D foundation
models}~\cite{triposr, instantmesh, lrm, crm} train on 3D banks, so defects
propagate into the learned prior. This paper evaluates the first consumer directly; for the second we report the dataset properties a training pipeline would rely on, but no training experiment.

We argue that for downstream consumption, \textbf{spatial and semantic
alignment} matters as much as raw scale. \emph{Spatial alignment}: a shared
coordinate frame with metric scale, axis convention, and category-appropriate
origin. \emph{Semantic alignment}: text, image, and geometry that genuinely
describe the same object, verifiable by cross-modal similarity. AmaraSpatial-10K
provides 10{,}071 assets that are simultaneously metric-scaled, semantically
anchored, PBR-ready, collision-aware, and richly annotated.

\paragraph{Claims and validation.}
We validate spatial and semantic properties through intrinsic audits
(\S\ref{sec:intrinsic}), two checks that do not depend on an LLM or VLM (calibration of the scale intervals against measured ABO dimensions and a three-rater human audit of 314 assets, \S\ref{sec:external_validation}), and three downstream studies: text-to-asset retrieval
(\S\ref{sec:benchmark_retrieval}), Holodeck scene composition
(\S\ref{sec:asset-usabilty}, 20 prompts $\times$ 2 seeds per pack with 95\% confidence intervals and a blinded human preference study), and Habitat-Sim physics stability
(\S\ref{sec:relevance-robotics}). The dataset is synthetic; we do not claim
parity with photogrammetric datasets such as GSO for material fidelity. The evidence supports standardised ingestion, static placement, and basic physics compatibility. We do not evaluate grasping, stacking, or other contact-rich manipulation, and we do not train an image-to-3D model on the dataset; both remain open (\S\ref{sec:discussion}).

\noindent\textbf{Contributions.}
\begin{enumerate}
    \item A public dataset of 10{,}071 spatially and semantically aligned 3D
    assets, each shipping with a metric-scaled \texttt{.glb}, PBR maps, a
    convex collision hull, a paired reference image, and multi-sentence text
    metadata (\S\ref{sec:dataset}).
    \item A reusable evaluation suite for 3D asset banks (SPS, Concept
    Density, anchor error, and cross-modal CLIP coherence), applied across
    AmaraSpatial-10K, Objaverse, HSSD, ABO, and GSO (\S\ref{sec:intrinsic}), with the LLM-derived scale intervals calibrated against measured object dimensions and the annotations audited by human raters (\S\ref{sec:external_validation}).
    \item Downstream evidence on three pipelines: $3.4\times$ CLIP Recall@5
    over Objaverse in retrieval (\S\ref{sec:benchmark_retrieval});
    better planned-scale matching, far less anchor remediation at ingest, and a blinded human preference in Holodeck scene composition
    (\S\ref{sec:asset-usabilty}); and $99.1\%$ settling stability with
    $\sim$$20\times$ lower per-load wall time under Habitat-Sim
    (\S\ref{sec:relevance-robotics}).
\end{enumerate}
\section{Related Work}
\label{sec:related}

\paragraph{3D Datasets.}
ShapeNet~\cite{shapenet} lacks PBR materials and real-world scale.
Objaverse~\cite{objaverse,objaversexl} provides 800K+ to 10M+ objects but with
highly variable quality: many meshes are non-manifold, mislabeled, or
arbitrarily scaled. GSO~\cite{gso} offers metric-scale scans but only
$\sim$1{,}000 assets. ABO~\cite{abo} and HSSD~\cite{hssd} contribute product
metadata and indoor-scene assets respectively, but neither pairs metric scale
and PBR materials with rich, retrieval-ready descriptions
(Table~\ref{tab:dataset_comparison}).

\paragraph{Downstream consumers.}
Embodied AI simulators (Habitat~\cite{habitat}, iGibson~\cite{igibson},
ProcTHOR~\cite{procthor}) and LLM-driven scene composition systems
(Holodeck~\cite{holodeck}, LayoutGPT~\cite{layoutgpt}) require per-object
metric scale, anchoring, and collision properties; inconsistent scale produces
physically implausible arrangements. CLIP-based text-to-3D
retrieval~\cite{clip} is bounded by description quality, and Objaverse's short
titles and generic tags limit conditioning precision.

\paragraph{Filtering vs.\ authoring.}
A natural objection is that Objaverse can be filtered into compliance. We test
this directly in \S\ref{sec:asset-usabilty}: an \emph{Objaverse Matched} pack
applying our full preprocessing pipeline to raw Objaverse assets reaches parity with AmaraSpatial-10K on scene-level geometric metrics under identical Holodeck settings once the pipeline has repaired it. The difference lies in that repair: only 19.0\% of Objaverse Matched assets arrive floor-anchored at ingest versus 89.9\% for AmaraSpatial-10K, and planned-scale matching remains better for AmaraSpatial-10K. Recent canonicalisation efforts~\cite{orientationmatters,canoverse} address orientation for Objaverse but not metric scale, anchors, or collision geometry.

\section{The AmaraSpatial-10K Dataset}
\label{sec:dataset}

\subsection{Overview and Access}
AmaraSpatial-10K contains 10{,}071 synthetic 3D assets across 10 top-level
categories and 476 subcategories. Indoor Scenes accounts for $\sim$26\% of
the collection, followed by Characters \& Creatures and City \& Transport.
The remaining themes cover long-tail domains (Nature \& Landscape, Sci-Fi \&
Cosmic, Food \& Beverage). Each asset ships as an optimised \texttt{.glb}
with embedded PBR materials, a paired 2D reference image, a convex collision
hull, and structured metadata (taxonomy, metric dimensions, anchor type,
forward axis, multi-sentence description) sufficient for downstream pipelines
to consume without additional preprocessing. The full per-subcategory
distribution is given in Appendix~\ref{app:taxonomy}; the generation
pipeline, spatial alignment, and curation protocol in
Appendix~\ref{app:generation_pipeline}.

The dataset is publicly available on Hugging
Face\footnote{\url{https://huggingface.co/datasets/ZeroOneCreative/amara-spatial-10k}}
under CC~BY~4.0, with a five-year maintenance commitment; code for the evaluation suite and scene generation is on GitHub\footnote{\url{https://github.com/01C-Amara/Amara-spatial-dataset}}. No canonical
train/val/test split is provided; consumers training on the dataset should
hold out a subset stratified by subcategory. Intended uses include
asset banks for scene composition and
robotics/embodied-AI simulation (ingestion, static placement, and basic physics), AR/VR prototyping, and use as a curated, pre-normalised candidate corpus for single-image-to-3D training, a use we have not yet evaluated experimentally. The dataset is not
intended for photorealistic product rendering or any deployment requiring
verified photogrammetric fidelity (e.g.\ LiDAR or depth-sensor simulation
benchmarks).

\subsection{Position Relative to Existing Datasets}
\label{sec:dataset_comparison_table}
Table~\ref{tab:dataset_comparison} compares AmaraSpatial-10K with prior 3D
datasets across the spatial and semantic properties required by downstream
consumers; an asset-level qualitative comparison across four representative
themes is provided in Appendix~\ref{app:qualitative_comparison}
(Figure~\ref{fig:qualitative_comparison}). A quantitative scale audit on the
Seating category follows in \S\ref{sec:intrinsic}.

\begin{table*}[ht]
\centering
\footnotesize
\setlength{\tabcolsep}{2.2pt}
\renewcommand{\arraystretch}{1.12}
\aboverulesep=0pt
\belowrulesep=0pt

\caption{\textbf{Dataset property comparison.} Properties required for downstream embodied-AI, game-development, and scene-composition pipelines. PBR: physically based rendering. Datasets: HSSD (Habitat Synthetic Scenes Dataset), ABO (Amazon Berkeley Objects), GSO (Google Scanned Objects). Shaded row is AmaraSpatial-10K.}
\label{tab:dataset_comparison}

\resizebox{\textwidth}{!}{%
\begin{tabular}{@{}l c *{7}{c}@{}}
\toprule
\textbf{Dataset} &
\textbf{Assets} &
\textbf{\makecell{Metric\\scale}} &
\textbf{\makecell{Correct\\anchors}} &
\textbf{\makecell{PBR\\materials}} &
\textbf{\makecell{Collision\\hulls}} &
\textbf{\makecell{Rich semantic\\descriptions}} &
\textbf{\makecell{Paired 2D\\images}} &
\textbf{\makecell{Consistent\\forward axis}} \\
\midrule
Objaverse~\cite{objaverse} & $\sim$800K & \xmark & \xmark & Partial & \xmark & Partial & Partial & \xmark \\
HSSD~\cite{hssd} & $\sim$12K & \cmark & \xmark & \xmark & \cmark & \xmark & \xmark & \cmark \\
ABO~\cite{abo} & $\sim$8K & \cmark & \xmark & Partial & \xmark & Partial & \cmark & \xmark \\
GSO~\cite{gso} & $\sim$1K & \cmark & \cmark & \cmark & \xmark & \xmark & \cmark & \xmark \\
\rowcolor{heavenlygold}
\textbf{AmaraSpatial-10K (Ours)} & \textbf{10K} & \cmark & \cmark & \cmark & \cmark & \cmark & \cmark & \cmark \\
\bottomrule
\end{tabular}}
\end{table*}
\section{An Evaluation Suite for 3D Asset Banks}
\label{sec:intrinsic}
We define a reusable suite of metrics that jointly assess whether an asset
bank is fit for downstream consumption: Scale Plausibility Score (SPS),
intra-category scale consistency, geometric and textural health, anchor
accuracy, collision-hull fidelity, cross-modal CLIP coherence, and 
Concept Density. Where possible we compute the same metrics on matched
subsets of Objaverse, HSSD, ABO, and GSO. Reference implementations are
released alongside the dataset. Four quantities depend on an LLM or VLM: the SPS plausible intervals (LLM-judged), the source dimension estimates used for scaling (LLM), forward-axis verification (VLM), and the original Holodeck perceptual scores (Gemini). Concept Density is keyword-bank matching, and CLIP coherence and all geometric, anchor, and collision metrics are measured, so none of these involves a judge. \S\ref{sec:external_validation} checks the LLM/VLM-dependent quantities against measured dimensions and human raters.

We extract all assets matching seating keywords (chair, armchair, sofa,
stool, couch) and measure their bounding-box heights, scoring against the
plausible interval $[0.6, 1.1]$\,m derived from the LLM-as-Judge protocol
(\S\ref{sec:sps_validation}). Objaverse's 181 matched seating assets span
0.02--115{,}276\,m with mean 717.79\,m; only 17.7\% fall within the
plausible range. AmaraSpatial-10K's 353 seating assets cluster at median
0.72\,m / mean 0.80\,m, with 56.7\% plausible (Table~\ref{tab:seating_scale} and histogram in Appendix~\ref{app:seating_distribution}).

\subsection{Scale Plausibility Score (SPS)}
\label{sec:scale_plausibility}

A binary in-range/out-of-range evaluation is too coarse: an asset 1\%
outside the plausible interval is penalised identically to one $10\times$
too large. We propose the \textbf{Scale Plausibility Score (SPS)}, a
continuous metric with full credit inside the expected range and smooth
proportional penalisation outside.

\paragraph{Definition.}
Let $x$ denote the measured primary-axis dimension (in m), $[\ell, u]$ the
LLM-judged plausible interval, and $h = (u-\ell)/2$ the half-width. With
boundary distance
$d(x,\ell,u) = \max(0, \ell-x) + \max(0, x-u)$,

\begin{equation}
    \text{SPS}(x, \ell, u) = \exp\!\left( -\left(d/h\right)^{2} \right).
    \label{eq:sps}
\end{equation}

SPS\,$=1.0$ for any $x \in [\ell, u]$, with Gaussian decay beyond. We
normalise by $h$ (not $u-\ell$) so that the transition band where SPS
decays from $1.0$ to $\approx 0.37$ is exactly one interval-width wide,
matching the intuition that ``one interval's worth of deviation'' is
meaningful. This puts narrow ranges (tea cup, $h{=}2.5$\,cm) and wide ones
(building, $h{=}48.5$\,m) on the same \emph{relative} scale: $d{=}h$
yields SPS\,$\approx 0.37$, $d{=}2h$ yields SPS\,$\approx 0.02$. Because $d$ is in metres and divided by the same category's $h$, every aggregated quantity is a unit-free ratio (rescaling all inputs to mm, cm, or km changes SPS by at most $4.4\times10^{-16}$). Rankings
are robust to the choice of decay function (Appendix~\ref{app:sps_sensitivity},
Kendall's $\tau\geq 0.94$). Inside $[\ell,u]$ SPS is 1 regardless of width, so absolute \% Perfect grows with interval breadth. No dataset gains from this, since the intervals are frozen before evaluation and applied identically to all datasets: scaling every interval width by $\times0.5$ to $\times3$ keeps the AmaraSpatial-10K minus Objaverse SPS gap positive throughout (0.25 to 0.48), and wider intervals help Objaverse more in relative terms. We illustrate the SPS curve in
Appendix~\ref{app:sps_curve} (Figure~\ref{fig:sps_curve}).

\paragraph{LLM-as-Judge protocol.} \label{sec:sps_validation}
Plausible intervals $[\ell,u]$ are generated by a \emph{separate} LLM
instance prompted only with the subcategory name, with no access to our
dataset's measured dimensions. We run three independent queries
($T{=}0.1$) and take the union of their intervals. Identical intervals
score every asset across all evaluated datasets without further
adjustment. Prompts and frozen intervals are released
(Appendix~\ref{app:llm_prompts}). Since the source dimensions used for scaling are also LLM estimates, agreement between the two could be circular; \S\ref{sec:external_validation} therefore calibrates the frozen intervals against physically measured object dimensions.

\begin{table*}[ht]
\centering
\scriptsize
\setlength{\tabcolsep}{2pt}
\renewcommand{\arraystretch}{1.08}
\aboverulesep=0pt
\belowrulesep=0pt
\caption{\textbf{Scale plausibility (SPS, Scale Plausibility Score) and intra-category consistency.} Amara rows cover the nine scale-audited categories; Objaverse matched columns use keyword-matched assets from the same categories. $[\ell,u]$ is the LLM-judged plausible height range; $\bar{x}$ is mean height in metres; CV (coefficient of variation) is $\sigma/\bar{x}$; \% Perfect is the share of assets with SPS\,$=1$.}
\label{tab:sps_results}
\label{tab:scale_cv}
\label{tab:scale_summary}
\resizebox{\textwidth}{!}{%
\begin{tabular}{@{}l c *{5}{>{\columncolor{heavenlygold}}r} r r r@{}}
\toprule
& & \multicolumn{5}{>{\columncolor{heavenlygold}}c}{\textbf{AmaraSpatial-10K}} & \multicolumn{3}{c}{\textbf{Objaverse matched}} \\
\textbf{Category} &
\textbf{$[\ell,u]$ (m)} &
\textbf{$N$} &
\textbf{$\bar{x}$ (m)} &
\textbf{CV $\downarrow$} &
\textbf{SPS $\uparrow$} &
\textbf{\% Perfect $\uparrow$} &
\textbf{$N$} &
\textbf{$\bar{x}$ (m)} &
\textbf{CV $\downarrow$} \\
\midrule
Architecture      & 3.0--100.0 & 733  & 12.73 & 2.39 & 0.988 & 38.9 & 629 & 6,667.04 & 1.34 \\
Vehicle           & 1.0--3.5   & 1101 & 4.68  & 4.14 & 0.762 & 32.0 & 553 & 799.57   & 8.02 \\
Animal            & 0.2--3.0   & 743  & 2.48  & 4.05 & 0.904 & 71.3 & 494 & 162.38   & 6.19 \\
Storage Furniture & 0.5--2.4   & 300  & 0.65  & 0.83 & 0.980 & 52.7 & 37  & 54.52    & 1.57 \\
Seating           & 0.6--1.1   & 353  & 0.93  & 1.03 & 0.812 & 56.7 & 175 & 739.42   & 11.75 \\
Table / Desk      & 0.4--0.9   & 558  & 1.14  & 1.98 & 0.672 & 44.4 & 301 & 237.15   & 7.54 \\
Electronics       & 0.05--0.9  & 207  & 0.99  & 1.55 & 0.768 & 64.7 & 141 & 69.43    & 3.64 \\
Tableware         & 0.05--0.30 & 589  & 0.93  & 2.17 & 0.479 & 32.4 & 109 & 8,724.42 & 10.13 \\
Nature (Flora)    & 0.1--20.0  & 638  & 3.75  & 4.20 & 0.981 & 95.0 & 417 & 1,979.25 & 10.14 \\
\midrule
\textbf{Overall}  & ---        & \textbf{5,222} & \textbf{3.89} & \textbf{3.40} & \textbf{0.815} & \textbf{51.8} & \textbf{2,856} & \textbf{1,723.18} & \textbf{9.92} \\
\bottomrule
\end{tabular}}
\vspace{2pt}
\begin{minipage}{0.98\textwidth}
\footnotesize \emph{Note.} The SPS aggregate covers the 5{,}222 AmaraSpatial-10K assets in the nine categories for which we define category-level height intervals and construct Objaverse keyword matches; the remaining 4{,}849 assets in the 10{,}071-asset release are outside this scale-audited subset. On the same intervals, the matched Objaverse aggregate has mean SPS $0.412$ and \% Perfect $7.7$.
\end{minipage}
\end{table*}

Across the nine evaluated categories, AmaraSpatial-10K achieves overall
mean SPS $0.815$ versus $0.412$ for the matched Objaverse subset, a
$1.98\times$ improvement (Table~\ref{tab:scale_summary}). Vehicle and
Tableware score lower (0.762, 0.479) not because of asset error but because
of category-level interval breadth: ``Vehicle'' subsumes bicycles
($\sim$1.6\,m) through trucks ($>$5\,m) but is scored against a single
$[1.0, 3.5]$\,m interval. At sub-category resolution, mean SPS rises to
$0.914$ (Vehicle) and $0.832$ (Tableware). Category-level numbers are
reported here for comparability with Objaverse, which lacks fine-grained
subcategory labels.

\subsection{Validation Independent of the LLM/VLM Pipeline}
\label{sec:external_validation}

\paragraph{Calibration against measured dimensions.}
The scaling step and the SPS intervals both draw on LLM size estimates, so high SPS alone could reflect two LLM calls agreeing with each other. We therefore score real objects against the same frozen intervals, using two independent sources of physically measured ABO~\cite{abo} heights: dimensions of ABO's metric 3D models and manufacturer specifications from its product metadata ($N\approx3{,}100$ each). Real heights fall inside our intervals 90.7\% and 88.7\% of the time (mean SPS 0.985 and 0.962; Table~\ref{tab:external_validation} in Appendix~\ref{app:external_validation}). ABO covers furniture and household goods, so this calibration holds for those categories; categories without a physical counterpart, such as fantasy creatures, cannot be calibrated this way.

\paragraph{Human audit.}
Three Zero One Creative employees who are not authors of this paper independently inspected a stratified sample of 314 assets for scale plausibility, forward axis, anchor placement, agreement between description and asset, material completeness, and major geometry defects that would block deployment. Pass rates range from 95.5\% (scale) to 100\% (materials), with no major geometry defects found (Table~\ref{tab:external_validation}).

\subsection{Per-Asset Quality}
\label{sec:per_asset_quality}

\paragraph{Intra-Category Scale Consistency.}
\label{sec:scale_consistency}

We measure how tightly assets within a category cluster in scale via the
coefficient of variation $\text{CV} = \sigma/\bar{x}$ of bounding-box
heights, computed on matched Objaverse subsets identified by keyword
search. The same scale summary in Table~\ref{tab:scale_summary} reports
mean CV $3.40$ for AmaraSpatial-10K versus $9.92$ for Objaverse, a
$2.9\times$ improvement. The contrast is starkest in Tableware ($2.17$
vs.\ $10.13$) and Seating ($1.03$ vs.\ $11.75$), where Objaverse spans
multiple orders of magnitude. Side-by-side box plots are in
Appendix~\ref{app:scale_boxplot}.

\paragraph{Geometric and Textural Health.}
\label{sec:geometric_health}

We audit all 10K meshes with \texttt{trimesh} / \texttt{PyMeshLab} and
compare against matched subsets from Objaverse, HSSD, and ABO
(Table~\ref{tab:geometric_health}). Under the
\texttt{trimesh.is\_watertight} definition, watertightness requires
edge-manifoldness and no boundary edges, so the two metrics coincide for
the audited subsets. AmaraSpatial-10K is competitive on manifoldness and
ships consistent UV coverage and 2048$\times$2048 textures. Face-count
distributions, including HSSD's $\sim$2-triangle spike at placeholder
geometry, are shown in Appendix~\ref{app:facecount}.

\paragraph{Spatial Alignment Verification.}
\label{sec:spatial_verification}

For bottom-anchored assets the origin should coincide with the
bounding-box bottom-centre; for centre-anchored assets, the centroid. We
report $\epsilon_{\text{anchor}}$, the Euclidean distance from mesh origin
to the semantically correct anchor (Table~\ref{tab:anchor_accuracy}). For
Objaverse, where pivots are essentially arbitrary, we report distance to
the nearest canonical anchor (charitable comparison, capped at 100\,m).
AmaraSpatial-10K achieves median $\epsilon_{\text{anchor}} = 1$\,mm, with
79.7\% of assets within 1\,cm of their semantic anchor; the corresponding
Objaverse median is 2.57\,m.

\paragraph{Collision hulls.}
\label{sec:collision_hull}

Each asset includes a convex collision proxy averaging 876 triangles
(95th pct.\ 2{,}458), with 99.99\% vertex containment and median volume
coverage $V_{\text{hull}}/V_{\text{bbox}} = 0.431$, a tighter physical
fit than HSSD's 0.200, the only baseline natively shipping hulls.

\subsection{Cross-Modal CLIP Coherence}
\label{sec:clip_coherence}

AmaraSpatial-10K is unique in providing three aligned modalities per asset, namely a text description, a 2D reference image, and a 3D mesh. We measure their internal consistency by computing pairwise CLIP cosine similarities across all three modalities, rendering each mesh from four canonical views ($+X$, $-X$, $+Y$, $-Y$) and averaging the resulting image embeddings. The three pairwise scores in Table~\ref{tab:clip_coherence} (Appendix~\ref{app:clip_distributions}) probe distinct steps of the generation pipeline: text to image, text to final mesh, and image to mesh.

The Reference Image~$\leftrightarrow$~3D Render coherence of 0.726 sits well above all text-to-visual scores, which isolates the text-to-image step rather than the 3D step as the dominant compositional drop. The 0.035-point gap between AmaraSpatial's and Objaverse's Text~$\leftrightarrow$~3D scores is a shift of the whole distribution, not a tail effect (Appendix~\ref{app:clip_distributions}, Figure~\ref{fig:clip_coherence}). The absolute Text~$\leftrightarrow$~3D cosine of 0.238 reflects the usual CLIP cosine range, not an encoder failure on our renders (Appendix~\ref{app:clip_lvis}), so the relative gap to Objaverse's 0.203 is the meaningful comparison.

\subsection{Semantic Description Richness}
\label{sec:semantic_richness}

We quantify textual metadata richness using two metrics. First, a \textbf{meaningful CLIP token count} measures raw descriptive length by tokenizing descriptions (CLIP ViT-L/14~\cite{clip}) and filtering out stopwords, non-alphabetical strings, and short tokens ($<3$ characters). Second, a novel \textbf{Concept Density} score (0--5) evaluates functional visual coverage by keyword matching, without an LLM judge. We define five visual constraint axes (\textbf{Color}, \textbf{Material}, \textbf{Style}, \textbf{Shape}, \textbf{Component}), each backed by a curated keyword bank, awarding one point per axis containing at least one matched keyword. While strict keyword matching is conservative (ignoring paraphrased concepts to act as a lower bound), it is applied uniformly to ensure fair relative comparisons across all datasets.

The semantic columns of Table~\ref{tab:semantic_richness} show that
AmaraSpatial-10K covers \textbf{2.62 of the 5 core visual constraint axes
per asset} on average, more than 18$\times$ Objaverse's coverage of 0.14.
Although Objaverse possesses the largest raw vocabulary (17,810 unique
tokens), this breadth arises from thousands of unique user-generated tags
rather than from coherent visual descriptions and does not close the
coverage gap. We expect the higher concept coverage to translate into
higher-precision conditioning for text-to-3D generative pipelines and
semantic retrieval systems, a hypothesis that the next section tests
directly.
\paragraph{Evaluation Suite Summary}
Taken together, Tables~\ref{tab:scale_summary}, \ref{tab:clip_coherence},
and \ref{tab:dashboard} show that AmaraSpatial-10K matches or exceeds
matched Objaverse on every intrinsic metric we can compute directly, with the smallest gap on mesh topology (watertight, manifold).

\begin{table*}[ht]
\centering
\scriptsize
\setlength{\tabcolsep}{1.2pt}
\renewcommand{\arraystretch}{1.12}
\aboverulesep=0pt
\belowrulesep=0pt
\caption{\textbf{Consolidated intrinsic quality dashboard.} Geometry, anchoring, and semantic metadata metrics are grouped by dataset. WT/manif.\ are watertight/manifold percentages; UV is the share of assets with UV texture coordinates; Tex.\ size is texture resolution in pixels; $\epsilon$ is anchor error; OOB is the out-of-bounding-box anchor rate; CD is concept density on a 0--5 scale. Arrows denote preferred direction. Objaverse anchor error is capped at 100\,m; dashes indicate metrics not reported in the current audit.}
\label{tab:geometric_health}
\label{tab:anchor_accuracy}
\label{tab:semantic_richness}
\label{tab:dashboard}
\resizebox{\textwidth}{!}{%
\begin{tabular}{@{}l l *{12}{c}@{}}
\toprule
\multicolumn{2}{c}{} & \multicolumn{5}{c}{\textbf{Geometry \& Textures}} & \multicolumn{4}{c}{\textbf{Anchoring}} & \multicolumn{3}{c}{\textbf{Semantic Metadata}} \\
\cmidrule(lr){3-7}\cmidrule(lr){8-11}\cmidrule(l){12-14}
\textbf{Dataset} &
\textbf{\makecell{Text\\source}} &
\textbf{\makecell{WT $\uparrow$\\(\%)}} &
\textbf{\makecell{Manif. $\uparrow$\\(\%)}} &
\textbf{\makecell{Mean\\faces}} &
\textbf{\makecell{UV $\uparrow$\\(\%)}} &
\textbf{\makecell{Tex.\\size}} &
\textbf{\makecell{Mean\\$\epsilon$ (m) $\downarrow$}} &
\textbf{\makecell{Med.\\(m) $\downarrow$}} &
\textbf{\makecell{OOB $\downarrow$\\(\%)}} &
\textbf{\makecell{$<$1cm $\uparrow$\\(\%)}} &
\textbf{\makecell{Mean\\tokens $\uparrow$}} &
\textbf{\makecell{Vocab\\size $\uparrow$}} &
\textbf{\makecell{CD $\uparrow$}} \\
\midrule
Objaverse & tags/titles & 59.8 & 59.8 & 148,569.6 & 94.4 & $\sim625^2$ & 23.974* & 2.569 & 35.2 & 4.2 & 20.2 & 17,810 & 0.14 \\
HSSD & tags & 54.4 & 54.4 & 10,917.2 & 79.7 & \makecell{prog.\\colors} & 0.169 & 0.049 & 27.0 & 25.1 & 3.3 & 199 & 0.01 \\
ABO & prod.\ desc. & 85.2 & 85.2 & 34,497.2 & 100.0 & $\sim3174^2$ & 0.087 & 0.056 & 16.7 & 29.4 & 36.7 & 2,977 & 1.01 \\
GSO & prod.\ names & --- & --- & --- & --- & --- & --- & --- & --- & --- & 8.6 & 220 & 0.54 \\
\rowcolor{heavenlygold}
\textbf{AmaraSpatial-10K} & \textbf{struct.\ desc.} & \textbf{61.7} & \textbf{61.7} & \textbf{47,038.5} & \textbf{100.0} & \textbf{$2048^2$} & \textbf{0.041} & \textbf{0.001} & \textbf{5.2} & \textbf{79.7} & \textbf{39.4} & \textbf{11,334} & \textbf{2.62} \\
\bottomrule
\end{tabular}}
\end{table*}

\section{Downstream Benchmark: Text-to-Asset Retrieval}
\label{sec:benchmark_retrieval}

We use this benchmark to test whether the description richness measured in \S\ref{sec:semantic_richness} translates into a measurable downstream gain.  Because the galleries carry different kinds of metadata, we define ground truth the same way for all of them through self-retrieval: each asset's own text metadata is the query and that asset is the single correct target. The gallery side contains only rendered images, so no text-to-text matching is possible. Each gallery asset is represented by the L2-normalised mean of CLIP ViT-L/14 image embeddings over its four orthographic renders, and we rank assets by cosine similarity between the CLIP text embedding of the query and the gallery embeddings. The same retrieval protocol is applied to AmaraSpatial-10K (multi-sentence descriptions), to a matched-size random sample of Objaverse (titles, descriptions, and tags), and to ABO~\cite{abo} (product-listing text). Our primary metric is \emph{CLIP Recall@5}, with R@1, R@10, R@25, and median retrieval rank reported alongside for robustness. The benchmark scores an asset and its metadata together: a high score requires text that singles out the asset and renders that CLIP can match to that text. It therefore mostly measures metadata quality, and the ablation below separates the two.

\begin{table}[ht]
\centering
\small
\setlength{\tabcolsep}{4pt}
\aboverulesep=0pt
\belowrulesep=0pt
\caption{\textbf{Semantic retrieval with description-richness ablation.} CLIP (Contrastive Language-Image Pre-training) Recall@$k$ (R@$k$, share of queries whose target is in the top $k$) from cosine similarity between ViT-L/14 text embeddings and asset embeddings (mean-pooled over four orthographic renders). $N$ is gallery size. The two AmaraSpatial-10K rows share gallery and renders, varying only the query text to isolate description length. The lower block compares ABO with a random AmaraSpatial-10K subset at the same gallery size ($N{=}3{,}000$); recall falls as $N$ grows, so only rows with equal $N$ are directly comparable. For this comparison we report R@5 only.}
\label{tab:retrieval_benchmark}
\resizebox{\linewidth}{!}{%
\begin{tabular}{l l c c c c c c}
\toprule
\textbf{Dataset} & \textbf{Query text} & \textbf{$N$} & \textbf{R@1 $\uparrow$} & \textbf{R@5 $\uparrow$} & \textbf{R@10 $\uparrow$} & \textbf{R@25 $\uparrow$} & \textbf{\makecell{Median \\ rank $\downarrow$}} \\
\midrule
\rowcolor{heavenlygold}
AmaraSpatial-10K & brief + full description   & 10{,}071 & 0.349 & 0.612 & 0.710 & 0.816 & 3 \\
AmaraSpatial-10K & brief description only     & 10{,}071 & 0.200 & 0.416 & 0.522 & 0.644 & 9 \\
Objaverse        & title + description + tags & 9{,}264  & 0.089 & 0.181 & 0.223 & 0.288 & 267 \\
\midrule
AmaraSpatial-10K & brief + full description & 3{,}000 & --- & 0.763 & --- & --- & --- \\
ABO              & product-listing text     & 3{,}000 & --- & 0.124 & --- & --- & --- \\
\bottomrule
\end{tabular}}
\end{table}

\noindent AmaraSpatial-10K's rich descriptions raise CLIP Recall@5 from 0.181 (Objaverse) to 0.612, a 3.4$\times$ improvement, and reduce the median retrieval rank from 267 to 3. The middle ablation row in Table~\ref{tab:retrieval_benchmark} isolates the contribution of description length on the same asset bank with identical renders. Stripping each query to the brief description alone roughly halves R@5 to 0.416, yet the result still exceeds Objaverse by 2.3$\times$, which indicates that the gap is driven by description content rather than by asset or render style. Figure~\ref{fig:retrieval_comparison} (Appendix~\ref{app:clip_lvis}) illustrates the failure mode qualitatively. All five Objaverse top-5 retrievals for \emph{``an ornate Victorian writing desk with brass handles''} carry only generic category tags (\emph{``furniture-home''}, \emph{``art-abstract''}, \emph{``architecture''}), whereas AmaraSpatial-10K returns descriptively named Victorian desks. Figure~\ref{fig:clip_classification_grid} (Appendix~\ref{app:clip_lvis}) confirms that the image encoder parses AmaraSpatial-10K renders correctly, so the gap reflects metadata quality rather than a CLIP failure mode. At a matched gallery of 3{,}000 assets under the identical protocol, ABO reaches R@5 $=0.124$ against $0.763$ for AmaraSpatial-10K, a $6.2\times$ gap, and ABO at $N{=}3{,}000$ still trails Objaverse at $N{=}9{,}264$ (0.181). This matches ABO's low Concept Density (1.01 vs.\ 2.62): it ships product-listing text, which rarely describes appearance. Re-captioning a gallery, for example with the VLM captions of TRELLIS-500K~\cite{trellis}, would raise its retrieval scores, but it cannot add metric scale, anchors, or collision geometry. We read the retrieval gain as evidence about metadata richness; evidence about 3D asset quality comes from \S\ref{sec:external_validation}, \S\ref{sec:asset-usabilty}, and \S\ref{sec:relevance-robotics}.\footnote{Maximum-cosine pooling over the four orthographic views instead of the mean yields CLIP R@5 $= 0.623$ for AmaraSpatial-10K and $0.196$ for matched Objaverse, confirming the gap is robust to the pooling strategy.}

\section{Asset Usability in Procedural Scene Composition}
\label{sec:asset-usabilty}

To test whether AmaraSpatial-10K's spatial invariants yield downstream gains, we use Holodeck~\cite{holodeck} (AI2-THOR/ProcThor~\cite{procthor} with GPT-4o~\cite{gpt4o} planning, CLIP/SBERT~\cite{clip,sbert} retrieval, constraint-based placement) and vary only the asset pack. \textbf{Objaverse Matched} is a 10{,}417-asset indoor subset of Objaverse~1.0 processed through our pipeline (controlled baseline). \textbf{AmaraSpatial-10K} is the experimental condition.\footnote{Both packs share an identical conversion and embedding pipeline (CLIP ViT-L-14, SBERT all-mpnet-base-v2, GLB-to-THOR with bounding-box colliders), so observed differences between them reflect asset source alone. Full pipeline details and the SPS-based filtering used to construct Objaverse Matched are in Appendix~\ref{app:pipeline}.} 
The submitted version of this study used six scenes per pack and had an error in our Objaverse GLB-to-THOR conversion that enlarged some Objaverse assets. We fixed the conversion, put both packs through identical preprocessing, embeddings, and collision representation, and generated scenes from 20 indoor prompts with two seeds each per pack (80 scenes, 40 paired comparisons). Geometric metrics come from the scene JSON (definitions in Table~\ref{tab:metrics_definitions}); we report paired differences with 95\% confidence intervals (Table~\ref{tab:quality_results}). The original six-scene study, which also included Holodeck's Default (AllenAI) pack and Gemini~2.5~Flash~\cite{gemini} perceptual scores, is kept for reference in Appendix~\ref{app:scene_composition}.

\begin{table}[ht]
\centering
\small
\setlength{\tabcolsep}{3.5pt}
\renewcommand{\arraystretch}{1.06}
\aboverulesep=0pt
\belowrulesep=0pt
\caption{\textbf{Scene quality across asset packs (corrected protocol).} 20 prompts $\times$ 2 seeds per pack; means over 40 scenes per pack and paired differences (AmaraSpatial-10K minus Objaverse Matched) with 95\% confidence intervals. \emph{Bottom:} blinded preference of three human raters over the 40 scene pairs; plausibility is rated on a 1--5 scale.}
\label{tab:quality_results}
\begin{tabular}{@{}l >{\columncolor{heavenlygold}}c c c@{}}
\toprule
\textbf{Metric} & \textbf{AmaraSpatial-10K} & \textbf{\makecell{Objaverse\\Matched}} & \textbf{Paired difference [95\% CI]} \\
\midrule
Object overlap $\downarrow$            & 0.125  & 0.000  & $+0.125$ [0.000, 0.401] \\
Floor contact (\%) $\uparrow$          & 100    & 100    & 0 \\
Scale consistency $\uparrow$           & 0.625  & 0.625  & 0 (same layout plan) \\
Spacing regularity $\uparrow$          & 0.559  & 0.581  & $-0.022$ [$-0.083$, 0.033] \\
Selected-vs-planned scale $\uparrow$   & \textbf{0.721} & 0.660 & $+0.061$ [0.033, 0.089] \\
Floor-anchored at ingest (\%) $\uparrow$ & \textbf{89.9} & 19.0 & --- \\
\midrule
Scene preferred (of 40)                & \textbf{27} & 13 & exact binomial $p=0.039$ \\
Physical plausibility (1--5) $\uparrow$ & \textbf{3.44} & 2.83 & $+0.617$ [0.233, 0.983] \\
\bottomrule
\end{tabular}
\end{table}

\paragraph{Geometric metrics.} Under the corrected protocol the scene-level geometric metrics are at parity: the overlap and spacing-regularity intervals include zero, both packs reach full floor contact, and scale consistency is identical because both packs replay the same layout plan. The overlap and floor-contact advantages reported in our submission came from the conversion error, and we withdraw them. Two differences remain. Assets retrieved from AmaraSpatial-10K match the dimensions in Holodeck's plan more closely ($+0.061$ [0.033, 0.089]), and 89.9\% of them arrive floor-anchored at ingest against 19.0\% for Objaverse Matched. Objaverse Matched reaches parity on floor contact only because our pipeline re-anchors it during ingestion.

\paragraph{Blinded human preference.} Three raters compared the 40 scene pairs with the asset pack hidden. They preferred the AmaraSpatial-10K scene in 27 of 40 pairs (exact binomial $p=0.039$) and rated its physical plausibility higher ($3.44$ vs.\ $2.83$; $+0.617$ [0.233, 0.983]). Agreement was moderate to substantial (Fleiss' $\kappa=0.609$~\cite{fleiss}), and 30 of the 40 pairs were unanimous.

\section{Relevance to Robotics and Embodied AI} \label{sec:relevance-robotics}
\noindent Beyond intrinsic spatial alignment (\S\ref{sec:intrinsic}), embodied-AI pipelines require assets to load and behave predictably under physics. This section tests static placement, settling, and collision compatibility. It does not test grasping, stacking, or contact-rich manipulation, where success depends jointly on the planner and the object, so a single-planner benchmark could not attribute failures to the asset. Raw synthetic meshes are often prone to unstable collisions and ground interpenetration, so we benchmark AmaraSpatial-10K's physical stability and tractability against Objaverse. Controlling for domain, we filter both datasets to realistically sized household objects ($\leq 2.0$\,m), yielding 2,569 AmaraSpatial-10K and 1,780 Objaverse assets. Each asset is released 10\,cm above a static floor for a 10-second Habitat-Sim~\cite{habitat} drop test (Bullet backend). AmaraSpatial-10K uses its pre-computed convex hulls and V-HACD decompositions; Objaverse uses Habitat-Sim's on-the-fly convex approximation over render meshes.

\textbf{Metrics.}~An asset passes if it settles (linear velocity $<$ 0.01 m/s, angular velocity $<$ 0.1 rad/s) without flying away (horizontal displacement $>$ 1.5 m) or excessive tunneling through the ground plane (interpenetration $<$ -0.05 m). We also measure wall-clock time for collision loading and physics steps, because geometrically stable assets that are computationally intractable remain impractical for large-scale embodied-AI training.

\textbf{Overall stability.}~Every asset in both subsets loaded and simulated (100\% load success). Table~\ref{tab:robotics_metrics_combined} shows that AmaraSpatial-10K reaches 99.1\% stability with V-HACD, with zero fly-aways or interpenetration failures; its single convex hull reaches 97.4\% (1 fly-away, 3 interpenetrations). Objaverse reaches 95.5\%, with 3 fly-aways and 63 interpenetrations. Because the AmaraSpatial-10K convex hull already outperforms the equivalent Objaverse baseline, the result suggests cleaner base geometry before multi-convex decomposition, while V-HACD further improves contact reliability for detailed assets.

\textbf{Continuous metrics.}~Ground penetration tightens from $-0.013 \pm 0.037$\,m (Objaverse) to $-0.010 \pm 0.006$\,m (Amara convex hull) and $-0.003 \pm 0.003$\,m (V-HACD), while per-load wall time drops from $0.442 \pm 1.105$\,s to $0.042 \pm 0.049$\,s and $0.022 \pm 0.024$\,s, roughly a $20\times$ speed-up. Horizontal displacement remains comparable (0.034--0.064\,m), reflecting normal settling dynamics rather than collision failure. Figure~\ref{fig:grasping_showcase} (Appendix~\ref{app:grasping}) shows qualitatively that the shipped hulls conform to the render mesh under gripper contact in MuJoCo; it is not a measure of grasp success.

\textbf{Preprocessing versus per-load cost.}~The per-load times above include Habitat-Sim's on-the-fly hull computation for Objaverse at every load, while AmaraSpatial-10K loads its shipped hulls. Table~\ref{tab:physics_cost} (Appendix~\ref{app:collision_hulls}) separates the one-time preprocessing cost. Computing a single convex hull takes about 7\,ms per asset (median; p95 21\,ms), so precomputing it pays off within the first load. V-HACD~\cite{vhacd} decomposition takes about 79\,s per asset at our released settings; we ship it precomputed, so users pay nothing, and a user who recomputed it would break even after about 189 loads.

\begin{table*}[ht]
\centering
\small
\aboverulesep=0pt
\belowrulesep=0pt
\setlength{\tabcolsep}{4pt}
\caption{\textbf{Dynamic physics and simulation tractability.} Stability pass rates for the Habitat-Sim drop test alongside continuous geometric and computational metrics (mean $\pm$ standard deviation). Flies: assets displaced more than 1.5\,m; Penetr.: assets sinking more than 5\,cm into the floor; XZ Disp.: horizontal displacement; Y Penetr.: deepest floor penetration; wall time is per load. Collision modes: $^1$internal Habitat-Sim convex hull, $^2$AmaraSpatial-10K convex hull, $^3$AmaraSpatial-10K V-HACD (Volumetric Hierarchical Approximate Convex Decomposition).}
\label{tab:robotics_metrics_combined}
\resizebox{\linewidth}{!}{%
\begin{tabular}{l ccc ccc}
\toprule
& \multicolumn{3}{c}{\textbf{Pass Rates \& Failures}} & \multicolumn{3}{c}{\textbf{Continuous Metrics \& Tractability}} \\
\cmidrule(lr){2-4} \cmidrule(lr){5-7}
\textbf{Dataset} & 
\textbf{\makecell{Stable~$\uparrow$}} & 
\textbf{\makecell{Flies~$\downarrow$}} & 
\textbf{\makecell{Penetr.~$\downarrow$}} & 
\textbf{\makecell{XZ Disp.~(m) $\downarrow$}} & 
\textbf{\makecell{Y Penetr.~(m) $\uparrow$}} & 
\textbf{\makecell{Wall Time~(s) $\downarrow$}} \\
\midrule
Objaverse$^1$ & 95.5\% & 3 & 63 & \textbf{0.034 $\pm$ 0.137} & -0.013 $\pm$ 0.037 & 0.442 $\pm$ 1.105 \\
AmaraSpatial-10K$^2$ & 97.4\% & 1 & 3 & 0.064 $\pm$ 0.167 & -0.010 $\pm$ 0.006 & 0.042 $\pm$ 0.049 \\
\rowcolor{heavenlygold}
\textbf{AmaraSpatial-10K$^3$} & \textbf{99.1\%} & \textbf{0} & \textbf{0} & 0.056 $\pm$ 0.162 & \textbf{-0.003 $\pm$ 0.003} & \textbf{0.022 $\pm$ 0.024} \\
\bottomrule
\end{tabular}}
\end{table*}

\section{Discussion}
\label{sec:discussion}
 
\subsection{Limitations}
 
\textbf{Dataset.} Assets are procedurally generated rather than scanned, which may introduce systematic biases in geometry and material accuracy relative to photogrammetric resources such as GSO. Coverage spans 10 top-level themes and 476 subcategories but is non-uniform: 23 subcategories contain only a single asset (Figure~\ref{fig:category_histogram}), adequate for taxonomic breadth but insufficient for subcategory-level generative training. The imbalance barely moves the headline metrics: excluding singleton subcategories changes mean SPS and intra-category CV by less than 0.001, and weighting all generation groups equally raises mean SPS to 0.832 (Objaverse: 0.412). Singletons are flagged in the metadata, and we advise stratifying by subcategory. PBR encoding is limited to Normal and Roughness maps, omitting full-BRDF phenomena such as subsurface scattering and anisotropy. Metric dimensions are LLM-estimated rather than physically measured; calibration against measured ABO dimensions and the human audit (\S\ref{sec:external_validation}) support them for household categories, but categories without a physical counterpart cannot be calibrated, and unusual or ambiguous objects remain edge cases. Textual metadata is English only.

\textbf{Evaluation.} Several evaluator-level caveats apply. Absolute SPS is bounded by category-level interval breadth: broad categories such as Vehicle score lower than tight ones such as Seating even when per-subcategory scale is plausible, so the constraint is in the evaluator's resolution rather than the asset. CLIP cross-modal scores should be read relatively because CLIP has known weaknesses in compositional reasoning, counting, and fine-grained attributes. The retrieval benchmark mainly measures metadata richness. The scene-composition study uses one generator, 20 prompts with two seeds, and three human raters; its geometric metrics are at parity, and the supported advantages are narrower (planned-scale matching, less anchor remediation, human preference).

\textbf{Scope of the embodied-AI claims.} Our physics evaluation covers loading, static placement, settling, and collision compatibility. It does not establish grasping, stacking, or manipulation performance, which needs a dedicated benchmark that controls for the planner; the grasping figure in Appendix~\ref{app:grasping} shows hull conformity only. We have not measured manual correction time or scale-fix rates in an end-to-end Isaac Sim or Unity ingestion workflow. We also have not trained an image-to-3D model on the dataset, so its value as a training corpus rests on its measured properties (consistent scale, complete PBR materials, rich captions) and remains to be tested. Future work covers a planner-controlled manipulation benchmark, a controlled Isaac Sim/Unity ingestion study, matched-size TripoSR/InstantMesh fine-tuning~\cite{triposr,instantmesh}, a LayoutGPT~\cite{layoutgpt} scene study, scaling to 100K assets, and a stand-alone Python package for the evaluation suite.
 
\subsection{Broader Impact}
AmaraSpatial-10K lowers the barrier for researchers without artist resources to assemble production-quality 3D asset banks and removes per-asset scale and anchor fixes for simulators that need accurate scale and PBR. Three risks merit acknowledgement. Synthetic asset banks concentrate the stylistic choices of their generating pipeline, and models trained exclusively on AmaraSpatial-10K may inherit those choices; future releases should diversify across pipelines. 3D generation at 10K scale carries a non-trivial compute footprint, so we discourage naive resynthesis where filtering existing datasets suffices. Finally, the dataset is released under CC~BY~4.0 for open research use; commercial deployers should consult the per-asset license metadata.

\section{Conclusion}
\label{sec:conclusion}
 
We presented AmaraSpatial-10K, a dataset of 10{,}071 synthetic 3D assets in which metric scale, semantic anchoring, PBR materials, collision hulls, and multi-sentence descriptions co-exist under a single coordinate convention. We accompanied it with a reusable evaluation suite (Scale Plausibility Score, intra-category scale consistency, anchor error, cross-modal CLIP coherence, and Concept Density), applied across AmaraSpatial-10K, Objaverse, HSSD, ABO, and GSO. Measured ABO dimensions and a three-rater audit of 314 assets check the LLM/VLM-derived annotations from outside the pipeline that produced them.

 CLIP text-to-asset retrieval improves by $3.4\times$ in Recall@5 over a matched Objaverse subset ($6.2\times$ over ABO at matched gallery size). In Holodeck scene composition with 40 paired scenes, geometric scene metrics are at parity with a matched Objaverse pack, while AmaraSpatial-10K matches planned object scale better, needs far less anchor remediation at ingest, and is preferred by blinded raters in 27 of 40 pairs. A Habitat-Sim drop test reaches $99.1\%$ settling stability with $\sim$$20\times$ lower per-load wall time. These results support standardised ingestion, static placement, and basic physics compatibility; manipulation and model training are untested. By publicly releasing the dataset alongside open-source implementations of the evaluation suite, we aim to make spatial and semantic alignment a default expectation for 3D asset banks.

\begin{ack}
\textbf{Funding.} This work was funded by Zero One Creative, where all authors were employed while the work was carried out. No third-party funding supported this work in the last 36 months.

\textbf{Competing interests.} At the time of the work, all authors were employees of Zero One Creative, which develops the in-house 3D generation engine used to create the assets (Appendix~\ref{app:generation_pipeline}). The human raters in \S\ref{sec:external_validation} and \S\ref{sec:asset-usabilty} were also Zero One Creative employees, though not authors.
\end{ack}

\bibliographystyle{plain}

\appendix

\newpage
\section{Full Category Taxonomy}
\label{app:taxonomy}

AmaraSpatial-10K spans 10 primary themes and 476 highly granular subcategories, totaling 10,071 curated assets. To provide a comprehensive overview while maintaining document conciseness, the complete taxonomy is flattened below. Each primary theme is listed with its total asset count, followed by its constituent subcategories.

\begin{figure}[ht]
    \centering
    \includegraphics[width=\linewidth]{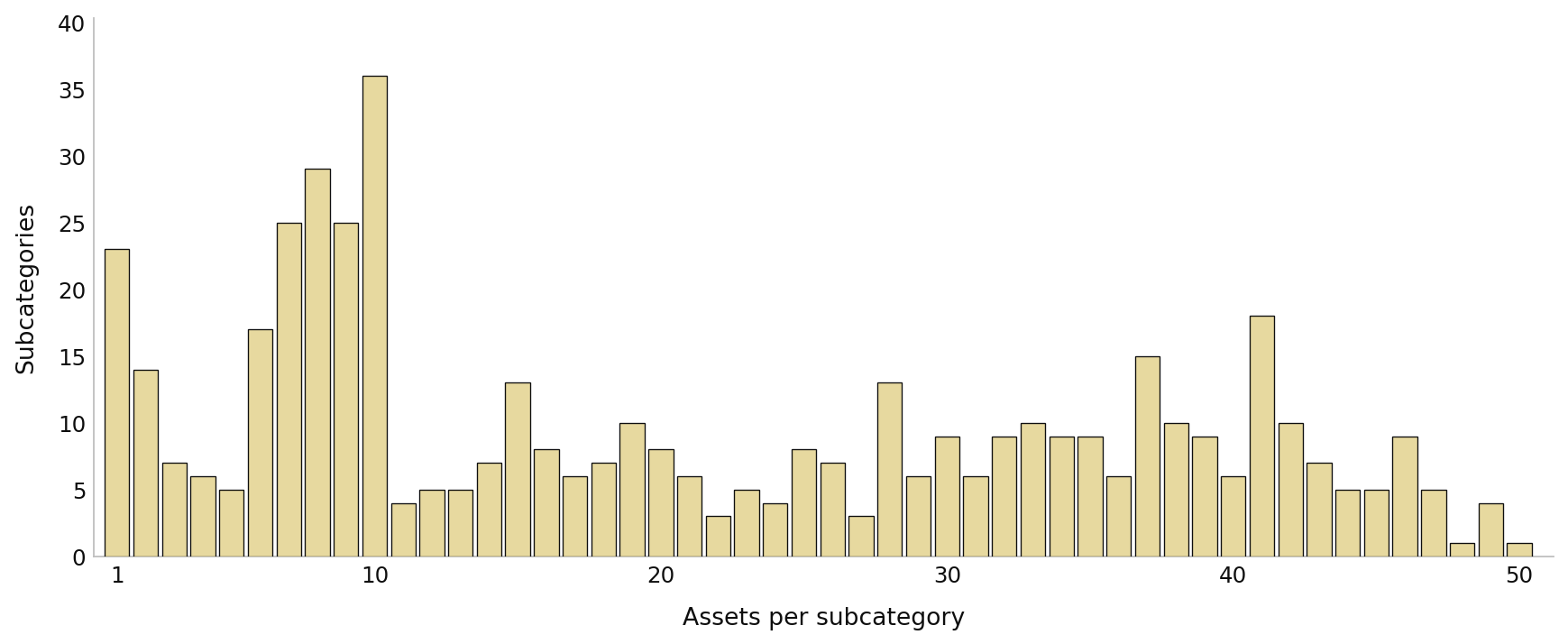}
    \caption{\textbf{Assets per Subcategory Distribution.} Subcategories are mostly populated with 5--15 assets each, with a heavy secondary cluster around 35--45 assets for visually rich categories (e.g.\ vehicles, architecture). 23 subcategories contain only a single asset each, retained for taxonomic breadth but not intended for subcategory-level learning, see \S\ref{sec:discussion} for discussion.}
    \label{fig:category_histogram}
\end{figure}
\FloatBarrier

\vspace{1em}

\begingroup
\small
\setlength{\parindent}{0pt}
\setlength{\parskip}{0.8em}

\textcolor{heavenlygold}{\textbf{Characters \& Creatures (1,749):}} 1930s Rubber Hose Style, AI Overseer, Android Citizen, Anthropomorphic Beast-Kin, Assault Droid, Assembly Line Robot, Atomic Age Retro-Futurists, Basilisk, Bear, Boar, Canary, Cargo Transport Robot, Cat, Cel-Shaded Anime Protagonists, Central AI Core, Chibi Super-Deformed, Chimera, Chunky Norse Stylized Warriors, Claymation-Style Figures, Combat Android, Construction Robot, Curious Child Robot, Cybernetic Mercenaries, Cyberpunk Neon Toons, Data Analysis Robot, Deer, Delivery Drone, Dog, Dragon, Drone Soldier, Eagle, Eldritch Cosmic Horrors, Engineer Robot, Ethereal Spirit Entities, Fairy Tale Storybook Illustrative, Fish, Flying Drone, Fox, Gecko, Geometric Abstract Humanoids, Glitch-Art Digital Entities, Griffin, Hamster, Heavy Loader Robot, Heavy Mech, Hedgehog, High-Fantasy Elemental Guardians, Hippogriff, Hive Mind AI, Iguana, Industrial Worker Robot, Inspection Drone, Kraken, Lab Assistant Robot, Low-Poly Retro Gaming Toons, Maintenance Robot, Mini Dragon, Mining Robot, Network Administrator AI, Old Rusty Robot, Owl, Papercraft and Origami Beings, Parrot, Patrol Robot, Pegasus, Phoenix, Pixar-esque Heroic Humanoids, Police Robot, Rabbit, Raccoon, Rebel Robot, Repair Drone, Road Maintenance Robot, Robot Bartender, Robot Cat, Robot Chef, Robot Dog, Robot Farmer, Robot Librarian, Robot Mayor, Robot Mechanic, Robot Taxi Driver, Robot Teacher, Robot Vendor, Scientist Robot, Scrap Collector Robot, Security Robot, Service Robot, Sewer Robot, Shield Robot, Snake, Sniper Drone, Soft-Body Plushie Characters, Steampunk Clockwork Automatons, Steampunk Victorian Inventors, Street Cleaning Robot, Stylized Urban Ninjas, Surveillance Drone, Tank Robot, Turtle, Unicorn, Urban Vinyl Art Toys, Victorian Gothic Stylization, Warehouse Robot, Watercolor Hand-Painted Avatars, Welding Robot, Wolf, Wyvern.

\textcolor{heavenlygold}{\textbf{Indoor Scenes (2,644):}} Ancient Museum Hall, Art Deco Boutique Lobby, Art Deco Casino Floor, Art Deco Hammam, Art Deco Maximalist, Asian Zen Waiting Room, Baroque Master Bedroom, Biophilic Greenhouse, Biophilic Greenhouse Study, Biophilic Jungle Eco-Resort, Biophilic Zen Sanctuary, Biophilic Zen Yoga Studio, Bohemian Creative Atelier, Bohemian Teen Bedroom, British Pub, Brutalist Museum Hall, Brutalist Outdoor Calisthenics Park, Brutalist Storage Warehouse, Contemporary Dentist Office, Contemporary TV Studio, Craftsman Workshop, Cyberpunk Command Center, Cyberpunk Garage, Cyberpunk High-Security Containment, Cyberpunk Modular Passageway, Cyberpunk Neon Coffee Kiosk, Dark Academia Sanctuary, Dystopian Industrial Service Tunnel, Edwardian Grand Hotel Suite, Food Futuristic Fast Food Restaurant, Food Retro 1950s Fast Food Restaurant, Futuristic Bio-Hacking Chamber, Futuristic Bio-Hacking Lab, Futuristic Capsule Hostel, Futuristic High-Tech Kitchen, Futuristic Museum Hall, Futuristic Operating Room, Futuristic Orbital Pod, Gothic Library, Gothic Revival Cloister, Gothic Wine Cellar, Himalayan Salt Meditation Cave, Hollywood Regency Casino Floor, Hotel Lobby, Industrial Loft Espresso Bar, Industrial Loft Hostel, Industrial Loft Lavatory, Industrial Loft Workspace, Industrial Professional Kitchen, Industrial Pub, Industrial Warehouse Boxing Club, Industrial Warehouse Loft, Industrial Workshop, Irish Pub, Japanese Zen Minimalist, Japanese Zen Minimalist Cafe, Kitchen Brutalist Commercial Kitchen, Kitchen Industrial Commercial Kitchen, Kitchen Minimalist Residential Kitchen, Kitchen Modern Farmhouse Residential Kitchen, Luxe Modern Lobby, Luxe Modern Master Bedroom, Maximalist Art Studio, Mediterranean Coastal Kitchen, Mediterranean Wine Cellar, Mid-Century Modern Hospital Examination Room, Mid-Century Modern Studio, Mid-Century Modern Wet Room, Minimalist Patient Room, Minimalist Zen Retreat, Modern Brutalist Penitentiary, Modern Pub, Neo-Classical Luxury Suite, Neoclassical Library, Neoclassical Museum Hall, Olympic-Scale Aquatic Pavilion, Retro 1980s Casino Floor, Room French Country Dining Room, Room Persian Dining Room, Room Traditional Classic European Dining Room, Rustic Farmhouse Kitchen, Rustic Wine Cellar, Scandinavian Hygge Nook, Scandinavian Hygge Retreat, Scandinavian Minimalist Kitchen, Steam Punk Pub, Steampunk Workshop, Traditional Classic European Library, Tropical Lobby, Tudor Pub, Vaporwave Music Recording Studio, Victorian Explorer’s Library, Victorian Gothic Dungeon, Victorian Greenhouse, Wabi-Sabi Master Bedroom, Wine Bar, Zen Thermal Onsen, Dining Area Art Deco Restaurant, Dining Area Mediterranean Restaurant, Dining Area Rustic Buffet Restaurant, Florist Retail Biophilic Florist Shop, Florist Retail French Country Bakery, Florist Retail Retro 1950s Delicatessen, Lab Mid-Century Modern Classroom, Lab Scandinavian Kindergarten Classroom, Studio Brutalist Studio Office, Toy Store Eclectic Toy Store, Toy Store Tudor Bookstore, Toy Store Vintage Bookstore, Room Mediterranean Sunroom, Room Victorian Grand Parlor.

\textcolor{heavenlygold}{\textbf{Furniture \& Household (762):}} Bed, Bench, Bohemian Hand-Woven Decor, Bookshelf, Cabinet, Chair, Coffee Table, Contemporary Smart Lighting, Desk, Dining Chair, Dining Table, Industrial Workshop Organizers, Kitchen Cabinet, Lamp, Minimalist Nordic Kitchenware, Mirror, Nightstand, Office Chair, Rug, Shelf, Sofa, Stool, TV Unit, Victorian Ornamental Hardware, Wardrobe.

\textcolor{heavenlygold}{\textbf{City \& Transport (1,402):}} Air Conditioning Units, Airplanes, Barriers, Benches, Bicycles, Bike Racks, Black Cabs, Boats, Bollards, Bus Stops, Cars, Cement Bags, Chimneys, City Buses, Construction Cones, Courtyard Sky Lounge, Crates, Cyberpunk Heavy Cargo Hauler, Deep-Sea Research Submersible, Delivery Motorcycle, Delivery Vans, Electrical Boxes, Fire Hydrants, Futuristic Urban VTOL, Garden Pavilion, Hatchback Cars, Ladders, London Office Buildings, London Benches and Bins, London Bridges and Tunnels, London Bus Shelters, London Double-Decker Buses, London Glass Buildings, London Historical Landmarks, London Industrial Buildings, London Market Stalls, London Modern Landmarks, London Pavements and Curbs, London Phone Booths, London Post Boxes, London Railways and Stations, London Religious Buildings, London Residential Buildings, London Restaurants Exteriors, London Shopfronts and Buildings, London Streetlights, London Traffic Lights, London Trees, London Underground, London Shrubs and Flower Beds, Lunar Multi-Terrain Rover, Luxury Car, Mailboxes, Motorcycles, Newspaper Stands, Pallets, Paris-Inspired Benches and Bins, Paris-Inspired Bridges and Tunnels, Paris-Inspired City Buses and Taxi, Paris-Inspired Emergency Vehicles, Paris-Inspired Glass Buildings, Paris-Inspired Historic-Style Buildings, Paris-Inspired Industrial Buildings, Paris-Inspired Mailboxes, Paris-Inspired Market Stalls, Paris-Inspired Metro System, Paris-Inspired Modern Landmark-Style Buildings, Paris-Inspired Office Buildings, Paris-Inspired Pavements and Curbs, Paris-Inspired Public Kiosks, Paris-Inspired Railways and Stations, Paris-Inspired Religious Buildings, Paris-Inspired Residential Buildings, Paris-Inspired Restaurant Exteriors, Paris-Inspired Shopfronts and Buildings, Paris-Inspired Streetlights, Paris-Inspired Trees, Parking Meters, Pipes, Post-Apocalyptic Scavenger Truck, Road Barriers, Road Blocks, Road Signs, Rooftop Equipment, Rooftop Terrace, SUV, Safety Barriers, Satellite Dishes, Scaffolding, Security Cameras, Sky Bar, Solar-Powered Hydrofoil Yacht, Speed Bumps, Steampunk Ironclad Locomotive, Street Lights and Lamp Posts, Street Signs, Taxi, Toolboxes, Traffic Cones, Traffic Lights, Trash Bins, Trucks, Ventilation Units, Village Square, Water Tanks, Wooden Planks, Yachts.

\textcolor{heavenlygold}{\textbf{Nature \& Landscape (1,088):}} Abandoned Ship Graveyard, Art Nouveau Cliff-Carved Alabaster Manors, Bioluminescent Hidden Grotto, Bushes, Cracked Ground, Dirt Piles, Fallen Logs, Flowers, Grass Clusters, Gravel, Ground Debris, Hearthside Apothecary Kitchen, Leaves, Moss, Moss-Covered Cobblestone Well, Mud Patches, Pebbles, Plants, Rocks and Boulders, Rugged Nordic Fjord Shore, Rustic Botanist's Greenhouse, Sand Piles, Sun-Drenched Mediterranean Cove, Tree Stumps, Trees, Tropical Resort Oasis, Vines, Vintage Attic Reading Nook.

\textcolor{heavenlygold}{\textbf{Sci-Fi \& Cosmic (620):}} Airlock and Docking Bay, Alien Planet Base, Alien Temple and Ruin, Cargo Hold and Storage Bay, Command Bridge and Cockpit, Cosmic Bar, Cryosleep Chamber, Lunar Base Interior Brutalist, Lunar Base Interior Futuristic, Mars Colony Habitat Early Settlement, Mars Colony Habitat Luxe Domed, Orbital Hotel Room, Planetarium Interior, Space Lounge, Space Observatory, Space Station Laboratory, Space Station Living Quarters, Subterranean Base, Underground Bunker, Clockwork Industrialism Machinery Modular Pipe-and-Valve Kits.

\textcolor{heavenlygold}{\textbf{History \& Culture (891):}} Bioluminescent Yggdrasil Miniature, Brutalist Totemic Concrete, Chinese Empire, Classical Neoclassical Marble, Coralline Galleon Ruins, Crystalline Ankh of Life, Egypt Civilisation, Epic Fantasy, Forged Iron Nordic Vegvisir, Gilded Solar Eye of Horus, Greece Empire, Hydraulic Recovery Platform, Industrial Salvage Exosuit, Iridescent Pearl Yin Yang, Kinetic Parametric Metal, Low-Poly Abstract Geometric, Marble Ouroboros Infinity, Medieval Castle, Medieval Tavern, Medieval Village, Monolithic Basalt Celtic Knot, Organic Surrealist Biomorphism, Persian Empire, Roman Empire, Sandstone Mayan Kalachakra, Shore-Grounded Modern Freighter, Slavic Style, Weathered Bronze Dharma Wheel.

\textcolor{heavenlygold}{\textbf{Fashion \& Clothing (432):}} Ancient Ceremonial Regalia, Avant-Garde Architectural Couture, Bio-Organic Jewelry, Biomorphic Organic Fashion, Bohemian Nomad Layering, Cybernetic Augmented Eyewear, Cybernetic Techwear, Cyberpunk Techwear, Ethereal Fantasy Silks, Futuristic Exoskeleton Footwear, High-End Leather Artistry, High-Fantasy Plate Armor, Interstellar EVA Suits, Luxury Chronograph Watches, Military Tactical Loadout, Modernized Samurai Armor, Post-Apocalyptic Scavenger Gear, Retro-Futurist Space Suits, Retro-Futuristic Flight Gear, Subaquatic Bio-Luminescent Suits, Tactical Military Loadouts, Urban Neo-Noir Formalwear, Victorian Steampunk Attire, Shoe Store Glam Sparkle Clothing Store.

\textcolor{heavenlygold}{\textbf{Food \& Beverage (225):}} Artisan Sourdough and Breads, Charcuterie and Aged Cheeses, Confectionery Candies, Exotic Tropical Fruits, Frozen Confections, Gourmet Patisserie, Hyper-Realistic Fast Food, Raw Earthy Root Vegetables, Sliced Citrus and Berries, Traditional Japanese Sushi.

\textcolor{heavenlygold}{\textbf{Music \& Play (258):}} Artisanal Resin Polyhedral Dice, Concert Woodwinds, Digital DJ Workstations, Futuristic Kinetic Sound Sculptures, Gothic Cathedral Pipe Organs, Hand-Carved Folk String Instruments, Hyper-Realistic Plush Textiles, Modern Electric Guitars, Modular Analog Synthesizers, Modular Cyberpunk Miniatures, Nordic Minimalist Wooden Toys, Orchestral Brass Section, Professional Studio Drum Kits, Vintage Grand Pianos, Vintage Tin Mechanicals.

\endgroup

\section{Annotation Validation Details}
\label{app:external_validation}

Table~\ref{tab:external_validation} gives the interval calibration and human-audit results summarised in \S\ref{sec:external_validation}.

\begin{table}[ht]
\centering
\footnotesize
\setlength{\tabcolsep}{4pt}
\aboverulesep=0pt
\belowrulesep=0pt
\caption{\textbf{Checks that do not use an LLM or VLM.} \emph{Top:} share of physically measured ABO (Amazon Berkeley Objects) heights that fall inside our frozen intervals, and their mean SPS (Scale Plausibility Score). \emph{Bottom:} stratified audit of 314 AmaraSpatial-10K assets by three non-author raters; observed pairwise agreement is 93--100\% across criteria.}
\label{tab:external_validation}
\begin{tabular}{@{}l c c@{}}
\toprule
\textbf{Interval calibration (ABO, $N\approx3{,}100$ each)} & \textbf{Inside interval $\uparrow$} & \textbf{Mean SPS $\uparrow$} \\
\midrule
Heights from metric 3D models   & 90.7\% & 0.985 \\
Manufacturer specifications     & 88.7\% & 0.962 \\
\midrule
\textbf{Human audit (314 assets, 3 raters)} & \multicolumn{2}{c}{\textbf{Rate}} \\
\midrule
Scale plausible $\uparrow$                 & \multicolumn{2}{c}{95.5\%} \\
Forward axis correct $\uparrow$            & \multicolumn{2}{c}{99.9\%} \\
Anchor correct $\uparrow$                  & \multicolumn{2}{c}{99.7\%} \\
Description matches asset $\uparrow$       & \multicolumn{2}{c}{98.4\%} \\
Materials complete $\uparrow$              & \multicolumn{2}{c}{100.0\%} \\
Major geometry defect $\downarrow$         & \multicolumn{2}{c}{0.0\%} \\
\bottomrule
\end{tabular}
\end{table}
\FloatBarrier

\section{CLIP Coherence Distributions}
\label{app:clip_distributions}

\begin{table}[ht]
\centering
\aboverulesep=0pt
\belowrulesep=0pt
\caption{\textbf{Cross-modal CLIP coherence.} Cosine similarity, mean $\pm$ standard deviation, $N{=}10{,}071$. Higher is better. For the Objaverse row, text concatenates the manifest's \texttt{name}, \texttt{description}, and \texttt{tags} (dropping assets with all three empty), and 3D renders use the same four canonical views.}
\label{tab:clip_coherence}
\begin{tabular}{l >{\columncolor{heavenlygold}}c c}
\toprule
\textbf{Modality Pair} & \textbf{AmaraSpatial-10K $\uparrow$} & \textbf{Objaverse $\uparrow$} \\
\midrule
Text $\leftrightarrow$ Ref.\ Image   & 0.303 $\pm$ 0.037 & N/A \\
Text $\leftrightarrow$ 3D Render      & 0.238 $\pm$ 0.041 & 0.203 $\pm$ 0.054 \\
Ref.\ Image $\leftrightarrow$ 3D Render & 0.726 $\pm$ 0.064 & N/A \\
\bottomrule
\end{tabular}
\end{table}

\begin{figure}[ht]
    \centering
    \includegraphics[width=\linewidth]{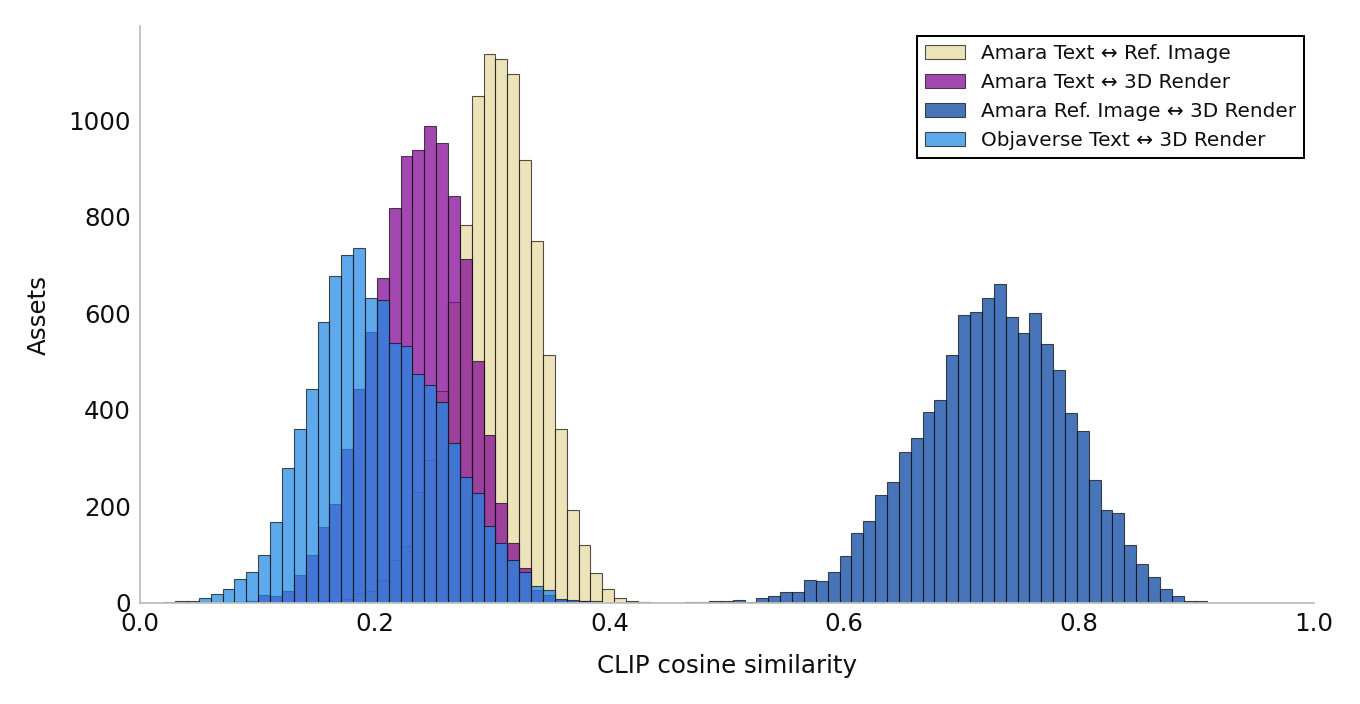}
    \caption{\textbf{CLIP coherence distribution.} Histograms of pairwise CLIP ViT-L/14 cosine similarity for AmaraSpatial-10K's three modality pairs (gold, purple, dark blue), alongside the matched Objaverse Text\,$\leftrightarrow$\,3D baseline (light blue). Distributional view of Table~\ref{tab:clip_coherence}.}
    \label{fig:clip_coherence}
\end{figure}

To confirm that the 0.035-point gap between AmaraSpatial-10K and Objaverse on Text~$\leftrightarrow$~3D coherence in Table~\ref{tab:clip_coherence} reflects a systematic shift rather than a tail-driven artefact, we plot the full pairwise CLIP cosine similarity distributions for each modality pair (Figure~\ref{fig:clip_coherence}).

\FloatBarrier

\section{CLIP LVIS Classification Verification}
\label{app:clip_lvis}

\begin{figure}[ht]
    \begin{minipage}{0.6\linewidth}
        \centering
        \includegraphics[width=\linewidth]{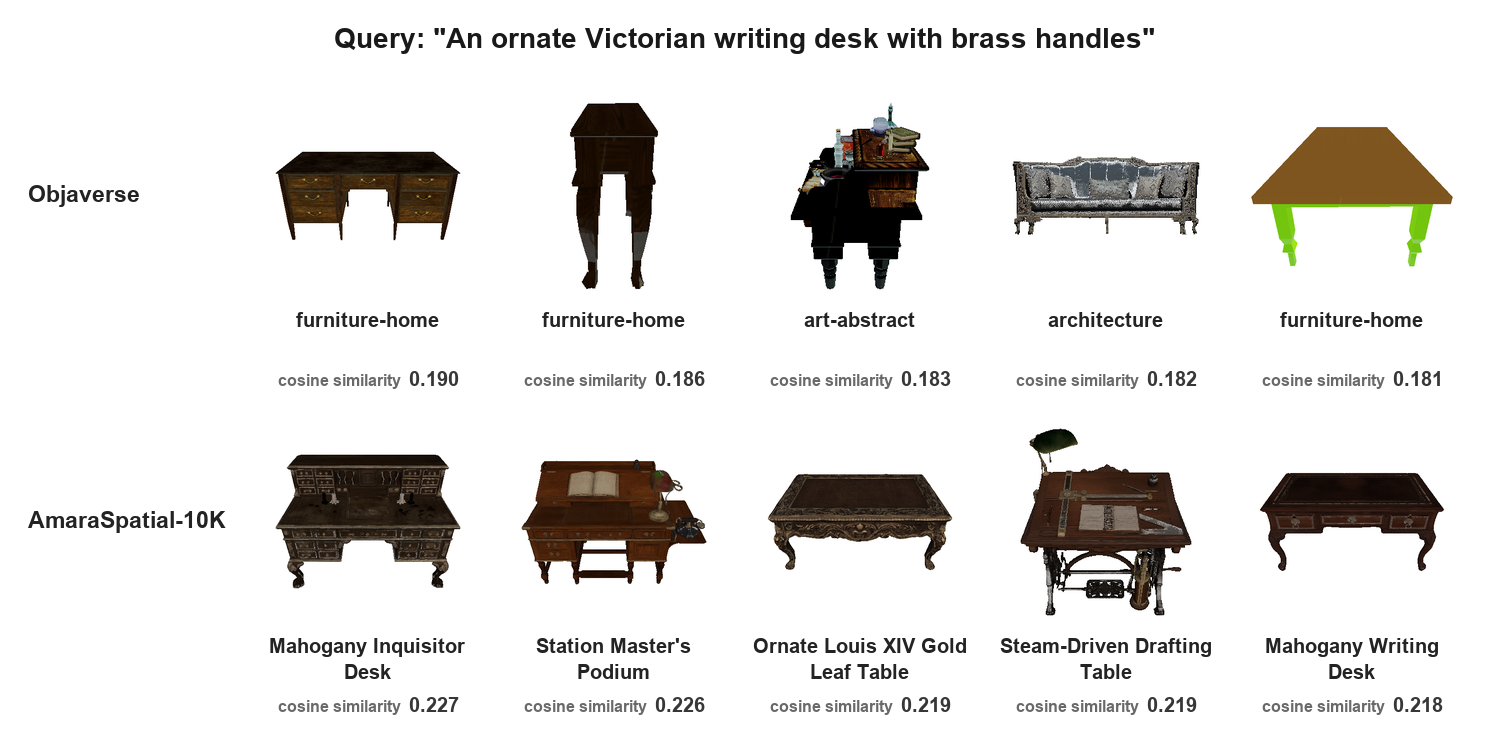}
    \end{minipage}\hfill
    \begin{minipage}{0.39\linewidth}
        \caption{\textbf{Qualitative CLIP retrieval.} Top-5 retrievals for \emph{``an ornate Victorian writing desk with brass handles''} from Objaverse (top) and AmaraSpatial-10K (bottom), ranked by CLIP ViT-L/14 cosine similarity (mean-pooled over four orthographic renders). Labels show each dataset's short descriptor (Objaverse \texttt{tags}, AmaraSpatial-10K \texttt{name}).}
        \label{fig:retrieval_comparison}
    \end{minipage}
\end{figure}

\begin{figure}[ht]
    \centering
    \includegraphics[width=\linewidth]{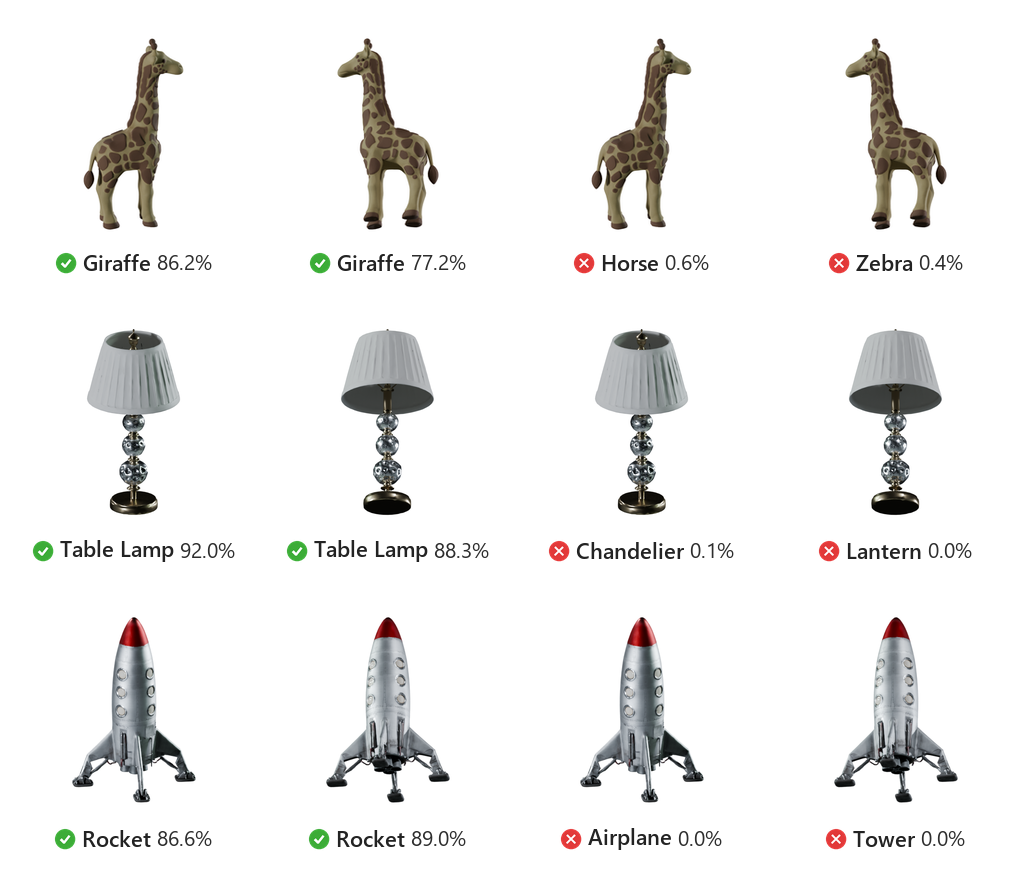}
    \caption{\textbf{CLIP classification of AmaraSpatial-10K renders.} For each asset we render two viewpoints and score them with CLIP (ViT-B/32, OpenAI pretraining) against a fixed LVIS vocabulary of 1{,}207 categories, following the Objaverse protocol (softmax over the full vocabulary). Columns 1--2 show the probability of the \emph{true} class under the two viewpoints, columns 3--4 re-display the same two renders scored against semantically close \emph{distractor} classes. The distractor probabilities are near zero across all three assets, confirming that the image encoder parses AmaraSpatial-10K renders correctly.}
    \label{fig:clip_classification_grid}
\end{figure}

To verify that the modest absolute Text~$\leftrightarrow$~3D coherence values reported in \S\ref{sec:clip_coherence} are not driven by encoder failure on AmaraSpatial-10K renders, we score each asset against a fixed LVIS vocabulary and compare the true-class probability against semantically close distractors (Figure~\ref{fig:clip_classification_grid}).

\FloatBarrier

\section{Asset-Level Qualitative Comparison}
\label{app:qualitative_comparison}

\begin{figure}[htbp]
    \centering
    \includegraphics[width=\linewidth]{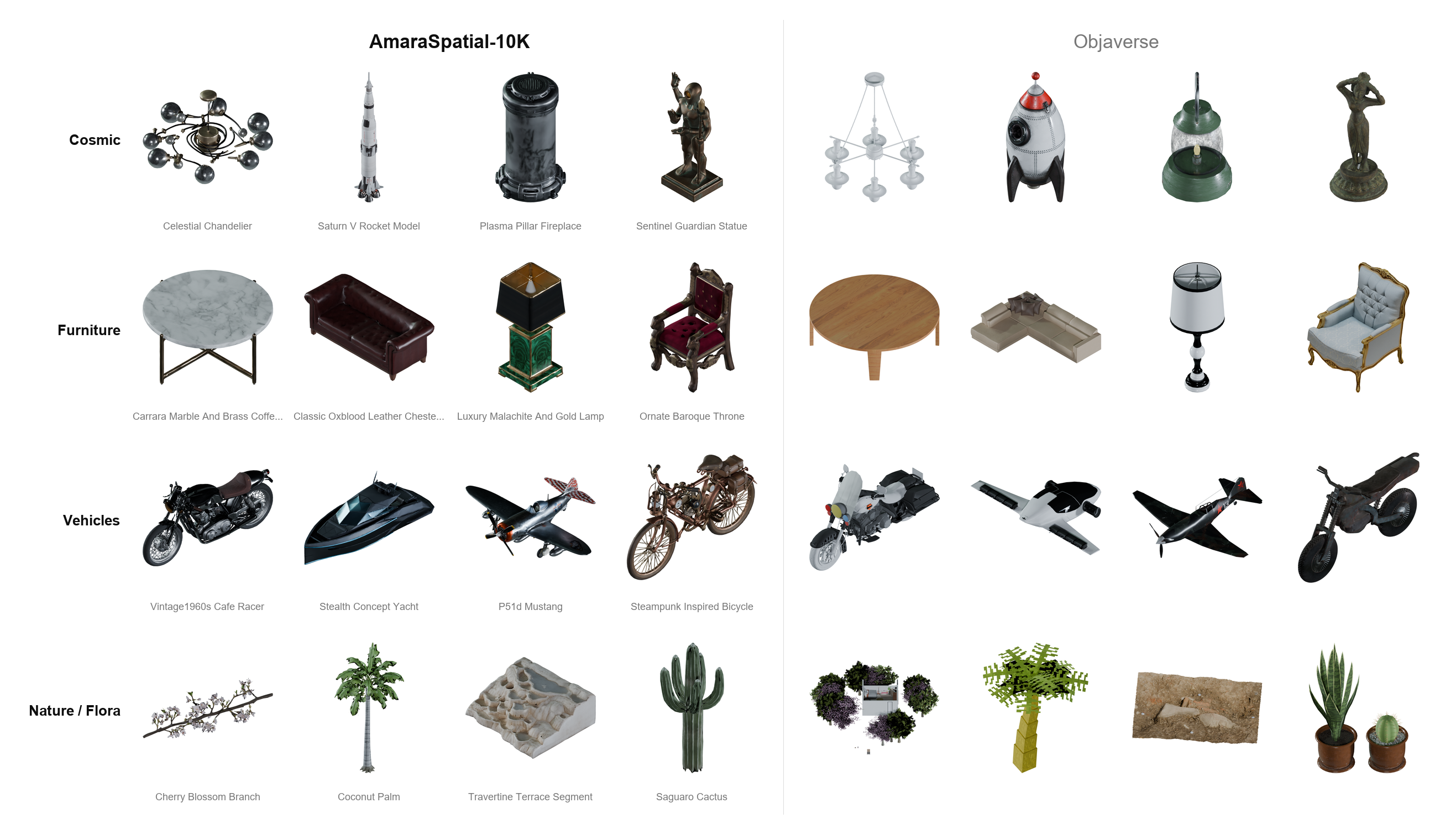}
    \caption{\textbf{Qualitative comparison across four representative
    themes.} Four assets per theme drawn from AmaraSpatial-10K (left) and
    Objaverse (right). AmaraSpatial-10K assets share consistent metric scale,
    canonical orientation, and PBR materials within a theme, whereas
    Objaverse assets, aggregated from heterogeneous creators, vary
    substantially in style, topology, and texture fidelity.}
    \label{fig:qualitative_comparison}
\end{figure}

Figure~\ref{fig:qualitative_comparison} provides an asset-level visual
comparison across four representative themes, complementing the
property-level comparison in Table~\ref{tab:dataset_comparison}.

\FloatBarrier

\section{Seating-Category Height Distribution}
\label{app:seating_distribution}

\begin{table}[ht]
\centering
\footnotesize
\setlength{\tabcolsep}{5.0pt}
\renewcommand{\arraystretch}{1.16}
\aboverulesep=0pt
\belowrulesep=0pt
\caption{\textbf{Seating-category scale comparison.} Bounding-box heights are scored against the plausible range $[0.6, 1.1]$\,m. $N$: number of matched seating assets; Plausible: share of assets inside the range. Objaverse has extreme outliers; its 5\% trimmed mean is 1.8\,m, still $\approx$3$\times$ the upper plausible bound.}
\label{tab:seating_scale}
\resizebox{\linewidth}{!}{%
\begin{tabular}{@{}l c r r r r c@{}}
\toprule
\textbf{Dataset} &
\textbf{$N$} &
\textbf{Median (m)} &
\textbf{Mean (m)} &
\textbf{Min (m)} &
\textbf{Max (m)} &
\textbf{Plausible (\%) $\uparrow$} \\
\midrule
Objaverse (Seating) & 181 & 2.44 & 717.79 & 0.020 & 115,276.9 & 17.7 \\
\rowcolor{heavenlygold}[\tabcolsep][\tabcolsep]
\textbf{AmaraSpatial-10K (Seating)} & \textbf{353} & \textbf{0.72} & \textbf{0.80} & \textbf{0.184} & \textbf{4.5} & \textbf{56.7} \\
\bottomrule
\end{tabular}}
\end{table}

\begin{figure}[htbp]
    \centering
    \includegraphics[width=\linewidth]{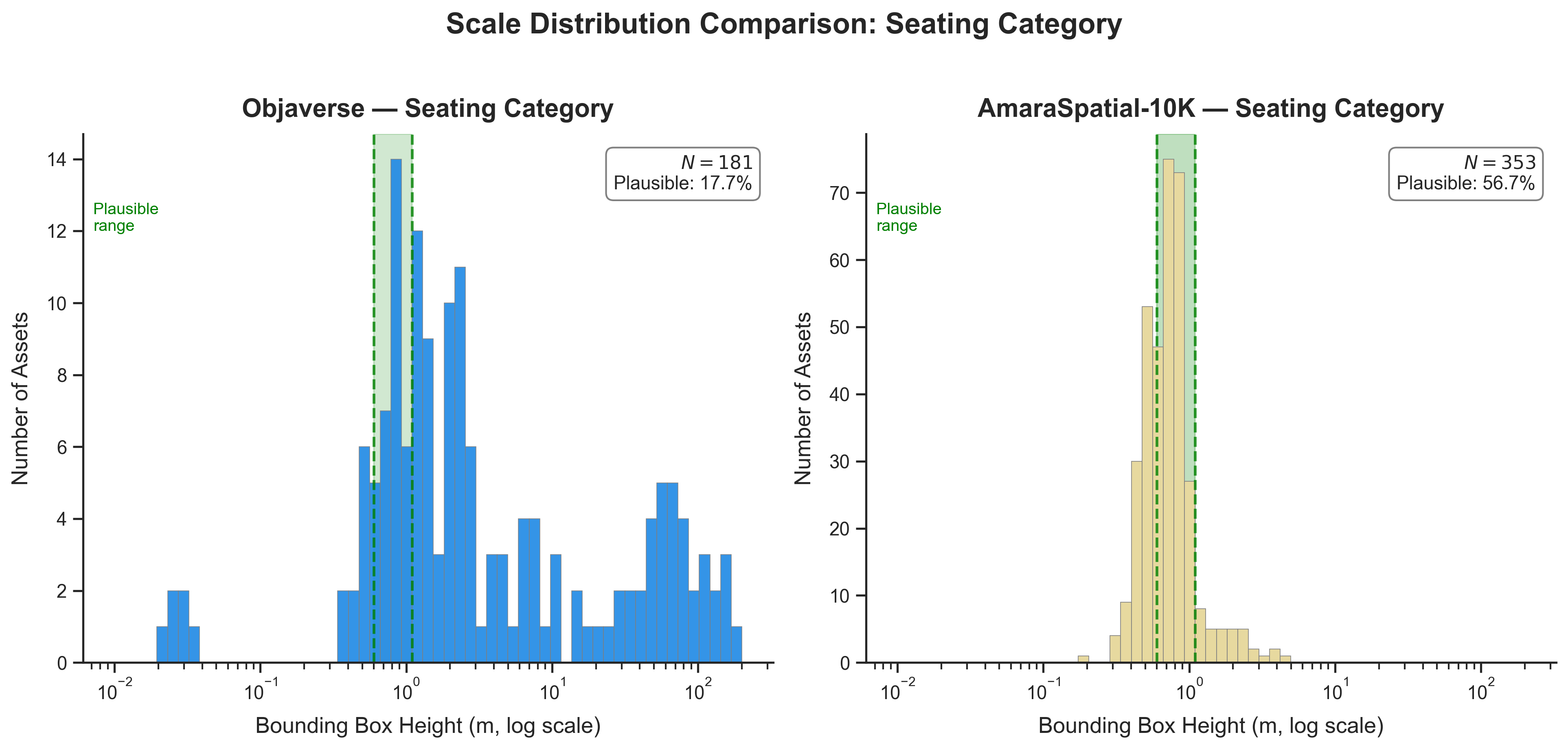}
    \caption{\textbf{Bounding box height distributions for the Seating
    category.} \textit{Left:} Objaverse ($N{=}181$) exhibits a multimodal,
    pathologically wide distribution spanning five orders of magnitude
    (0.02\,m to 115{,}276\,m). \textit{Right:} AmaraSpatial-10K
    ($N{=}353$) shows a tight, physically grounded distribution centred
    around a median of 0.72\,m. Both axes use a logarithmic scale to
    accommodate the dynamic range of Objaverse.}
    \label{fig:scale_dist}
\end{figure}

Figure~\ref{fig:scale_dist} visualises the bounding-box height
distributions summarised in Table~\ref{tab:seating_scale}.

\FloatBarrier

\section{Real-World Metric Scaling}
\label{app:metric_scaling}

\begin{figure}[htbp]
    \centering
    \includegraphics[width=\linewidth]{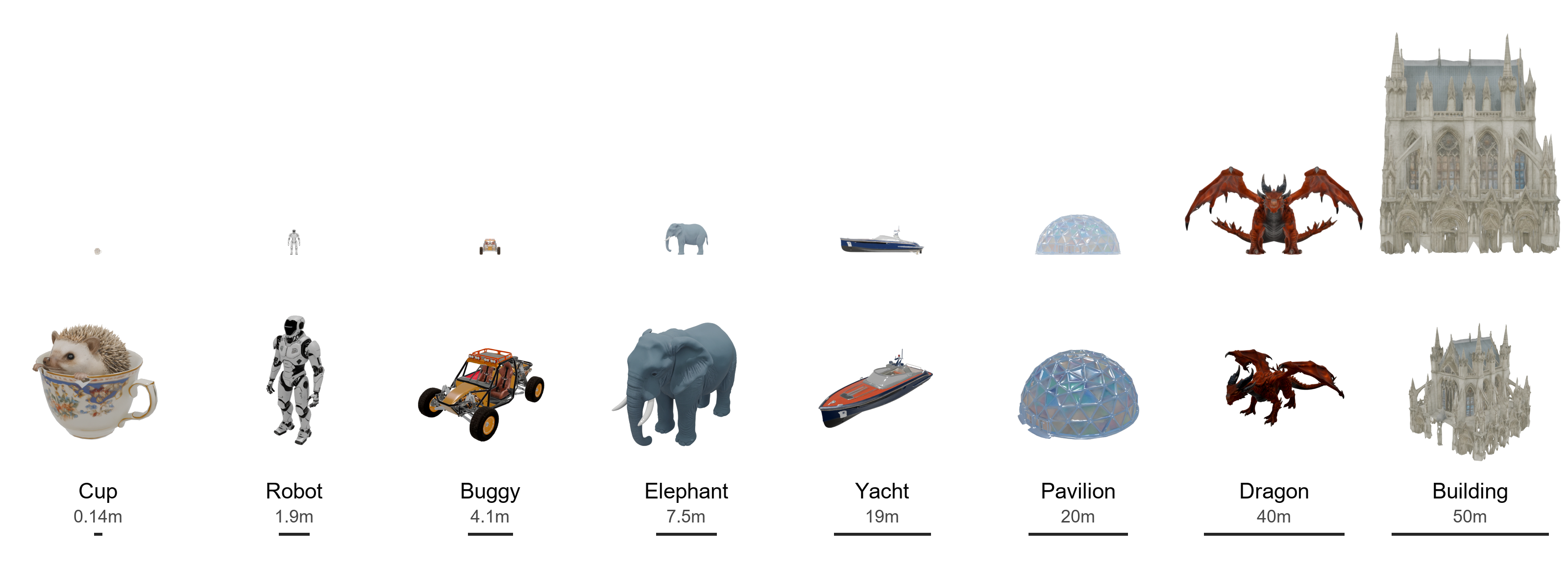}
    \caption{\textbf{Real-world metric scaling across AmaraSpatial-10K.}
    Eight representative assets rendered at shared ground scale, from cup
    to cathedral. All assets share a common metric ground truth, so no
    per-asset normalization is needed before placement. For fantasy
    creatures (e.g.\ dragon at 40\,m) the metric scale reflects design
    intent encoded in the asset's description and matches the LLM-judged
    plausible range for that subcategory, with no physical ground truth
    available for these cases.}
    \label{fig:metric_scaling}
\end{figure}

To make the metric-scale invariant claimed in \S\ref{sec:scale_plausibility} visually concrete, we render eight representative AmaraSpatial-10K assets at shared ground scale, spanning roughly four orders of magnitude from a tea cup to a cathedral (Figure~\ref{fig:metric_scaling}).

\FloatBarrier

\section{SPS Curve Visualisation}
\label{app:sps_curve}

\begin{figure}[htbp]
    \centering
    \includegraphics[width=\linewidth]{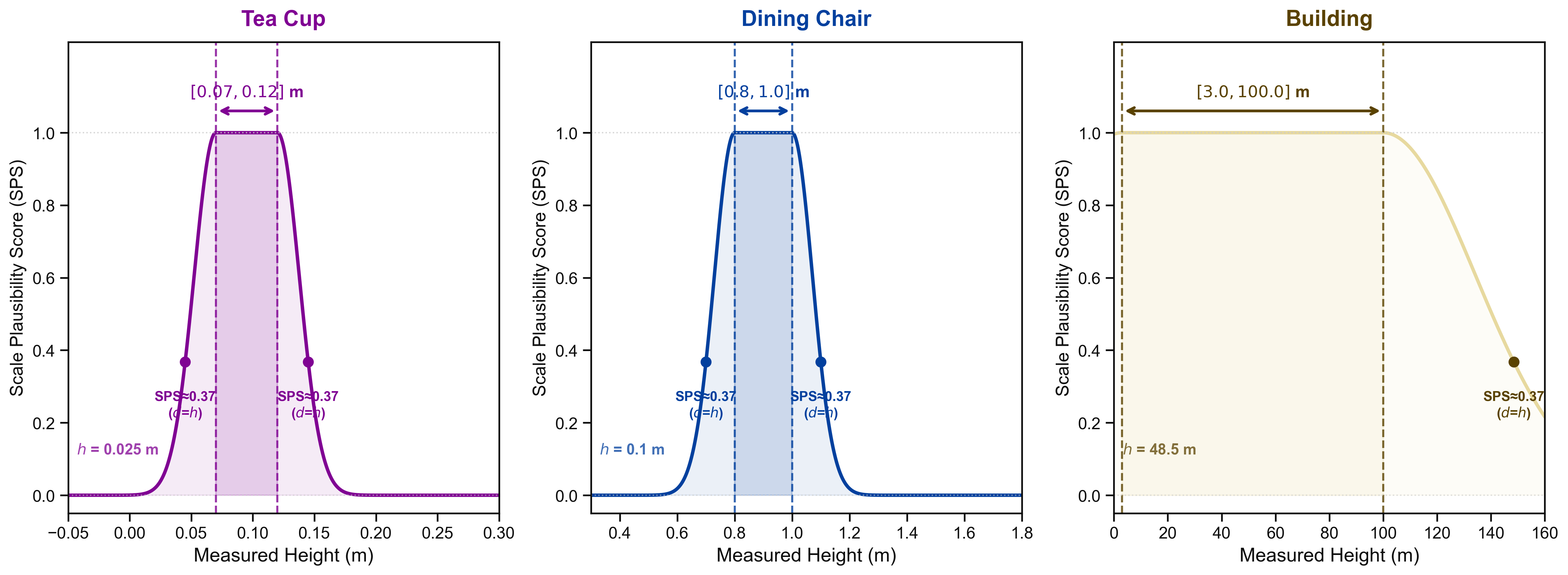}
    \caption{\textbf{Scale Plausibility Score (SPS) as a function of
    measured height for three representative subcategories.} Each panel
    shows the SPS curve Eq.~(\ref{eq:sps}) over the relevant measurement
    range. The shaded plateau ($\text{SPS}=1.0$) corresponds to the
    LLM-judged plausible interval $[\ell, u]$ (dashed vertical lines).
    Outside this interval, SPS decays symmetrically via a Gaussian with
    half-width $h=(u-\ell)/2$: an asset at distance $d=h$ from the nearest
    boundary scores $\approx 0.37$, and at $d=2h$ scores $\approx 0.02$.
    The normalization by $h$ ensures that a narrow-interval subcategory
    (Tea Cup, $h=2.5$\,cm) and a wide-interval subcategory (Building,
    $h=48.5$\,m) are penalized on the same \emph{relative} scale.
    \emph{Note:} this figure uses illustrative subcategory-level intervals
    (Tea Cup, Dining Chair, Building); Table~\ref{tab:sps_results} reports
    SPS against the broader category-level intervals (Tableware, Seating,
    Architecture).}
    \label{fig:sps_curve}
\end{figure}

To make the Scale Plausibility Score defined in Eq.~\ref{eq:sps} concrete, we plot the SPS curve for three subcategories spanning the dynamic range of plausible-interval widths in AmaraSpatial-10K, demonstrating how the half-width normalization places narrow-interval and wide-interval categories on a common penalty scale (Figure~\ref{fig:sps_curve}).

\FloatBarrier

\section{Intra-Category Scale Box Plots}
\label{app:scale_boxplot}

\begin{figure}[htbp]
    \centering
    \includegraphics[width=\linewidth]{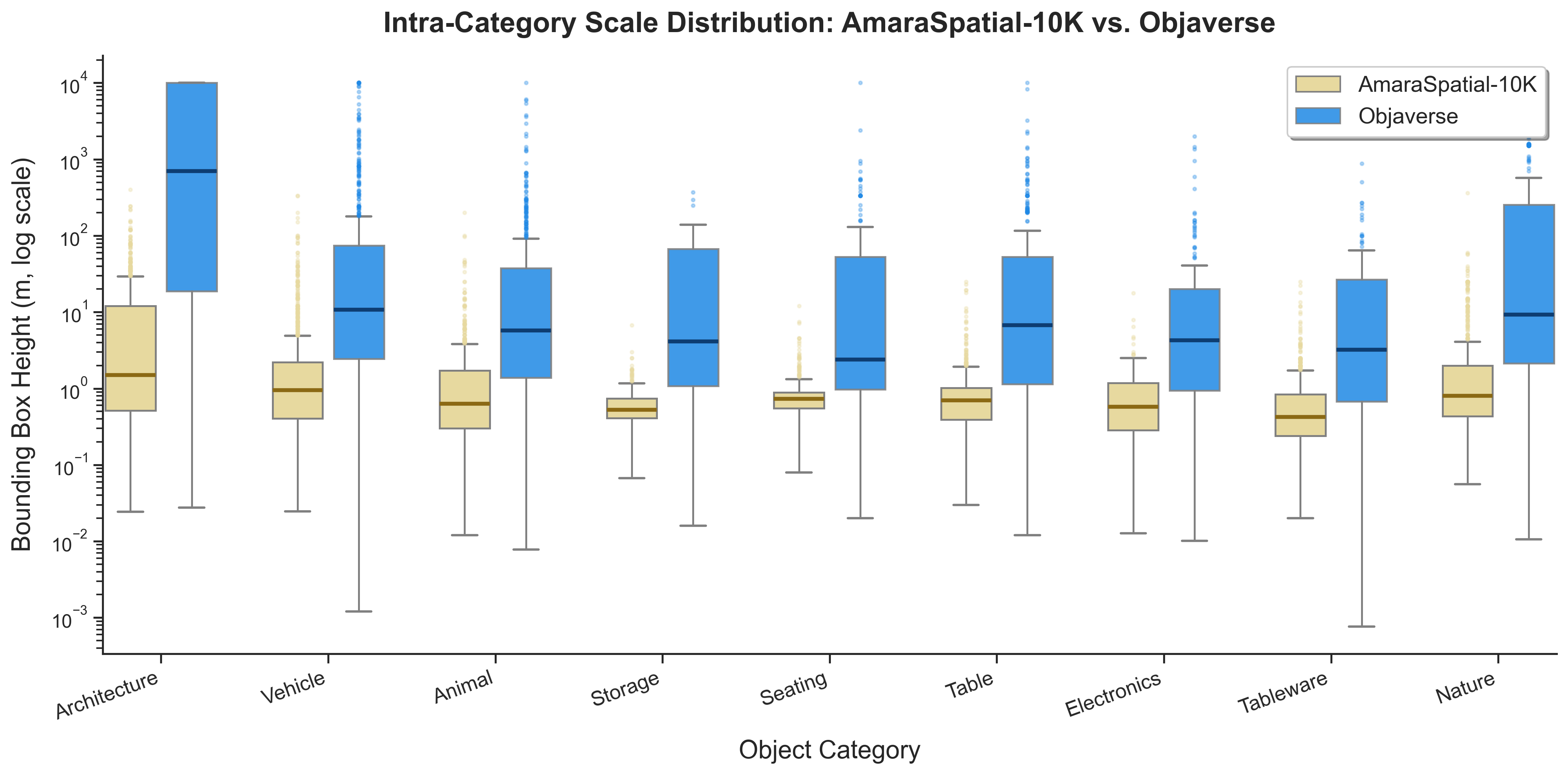}
    \caption{\textbf{Intra-category scale distribution.} Side-by-side box
    plots of bounding box height (log scale) for each object category.
    \textcolor[HTML]{E7D99F}{\rule{1em}{0.8em}}~AmaraSpatial-10K (Heavenly
    Gold ``Ours'') shows tight, physically plausible distributions centred
    around real-world object sizes.
    \textcolor[HTML]{1E88E5}{\rule{1em}{0.8em}}~Objaverse (Blue) exhibits
    dramatically wider boxes and extreme outliers spanning several orders
    of magnitude, confirming severe scale inconsistency across all
    categories.}
    \label{fig:scale_boxplot}
\end{figure}

Figure~\ref{fig:scale_boxplot} visualises the per-category bounding-box
height distributions summarised by the coefficient-of-variation values in
Table~\ref{tab:scale_cv}.

\FloatBarrier

\section{Face-Count Distributions}
\label{app:facecount}

\begin{figure}[htbp]
    \centering
    \includegraphics[width=0.8\linewidth]{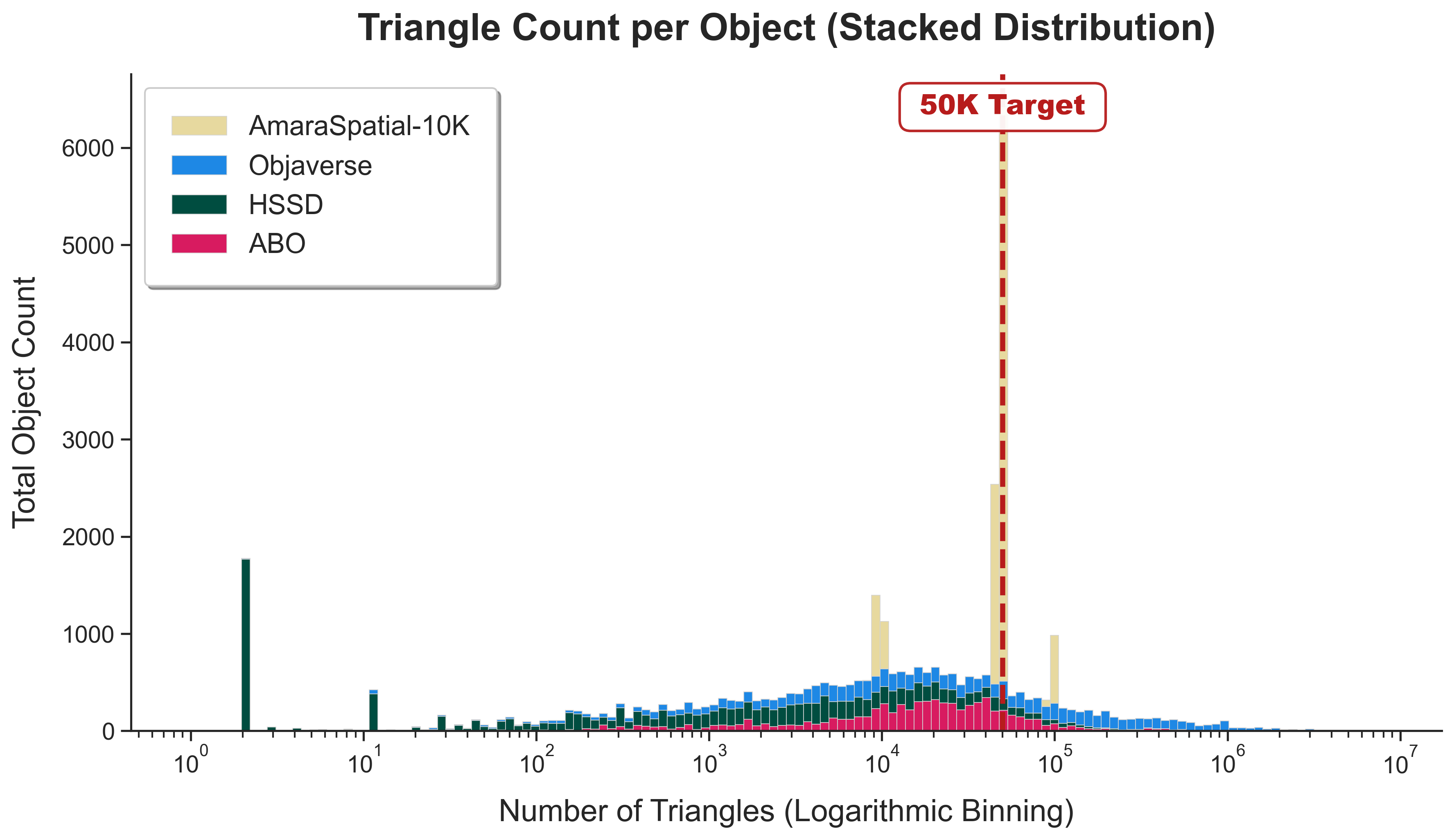}
    \caption{\textbf{Face count distributions across datasets.} The
    majority of AmaraSpatial-10K assets target $\sim$50K triangles. To
    support low-poly applications, roughly 2{,}000 assets are optimised to
    $\sim$10K triangles; $\sim$1{,}000 ``hero'' assets retain higher
    geometric detail at $\sim$100K. HSSD's distribution contains a visible
    spike at $\sim$2 triangles corresponding to placeholder/primitive
    geometry, absent in AmaraSpatial-10K.}
    \label{fig:face_count_histogram}
\end{figure}

To confirm that the polycount targets described in \S\ref{sec:geometric_health} are realised in practice, we plot the per-asset triangle-count distribution for AmaraSpatial-10K alongside HSSD, also exposing HSSD's spike at $\sim$2 triangles from placeholder geometry (Figure~\ref{fig:face_count_histogram}).

\FloatBarrier

\section{Collision Hull Statistics}
\label{app:collision_hulls}

\begin{table}[ht]
\centering
\small
\setlength{\tabcolsep}{4pt}
\aboverulesep=0pt
\belowrulesep=0pt
\caption{\textbf{One-time preprocessing versus per-load simulation cost} per asset. Per-load time is the mean wall time of Table~\ref{tab:robotics_metrics_combined}. Break-even is the number of loads after which precomputing is cheaper than Habitat-Sim's on-the-fly hull.}
\label{tab:physics_cost}
\resizebox{\linewidth}{!}{%
\begin{tabular}{@{}l c c c@{}}
\toprule
\textbf{Collision mode} & \textbf{Preprocessing (one-time)} & \textbf{Per load (s) $\downarrow$} & \textbf{Break-even} \\
\midrule
Objaverse, Habitat-Sim internal hull & none (paid at each load) & 0.442 & --- \\
AmaraSpatial-10K convex hull & $\sim$7\,ms median (p95 21\,ms) & 0.042 & $<1$ load \\
AmaraSpatial-10K V-HACD & $\sim$79\,s, shipped precomputed & 0.022 & $\sim$189 loads if recomputed \\
\bottomrule
\end{tabular}}
\end{table}

Table~\ref{tab:collision_hulls} reports the full collision-hull
statistics summarised in \S\ref{sec:collision_hull}.

\begin{table}[ht]
\centering
\caption{Collision hull statistics across all AmaraSpatial-10K assets.}
\label{tab:collision_hulls}
\begin{tabular}{lc}
\toprule
\textbf{Metric} & \textbf{Value} \\
\midrule
Mean hull triangles                     & 876.6 \\
95th percentile hull triangles          & 2{,}458 \\
Median volume coverage ($V_{\text{hull}} / V_{\text{bbox}}$) & 0.431 \\
Vertex containment (\%)                 & 99.99 \\
\bottomrule
\end{tabular}
\end{table}
\FloatBarrier

\section[Collision-Hull Contact Conformity]{Collision-Hull Contact Conformity}
\label{app:grasping}

\begin{figure}[ht]
    \centering
    \includegraphics[width=\linewidth]{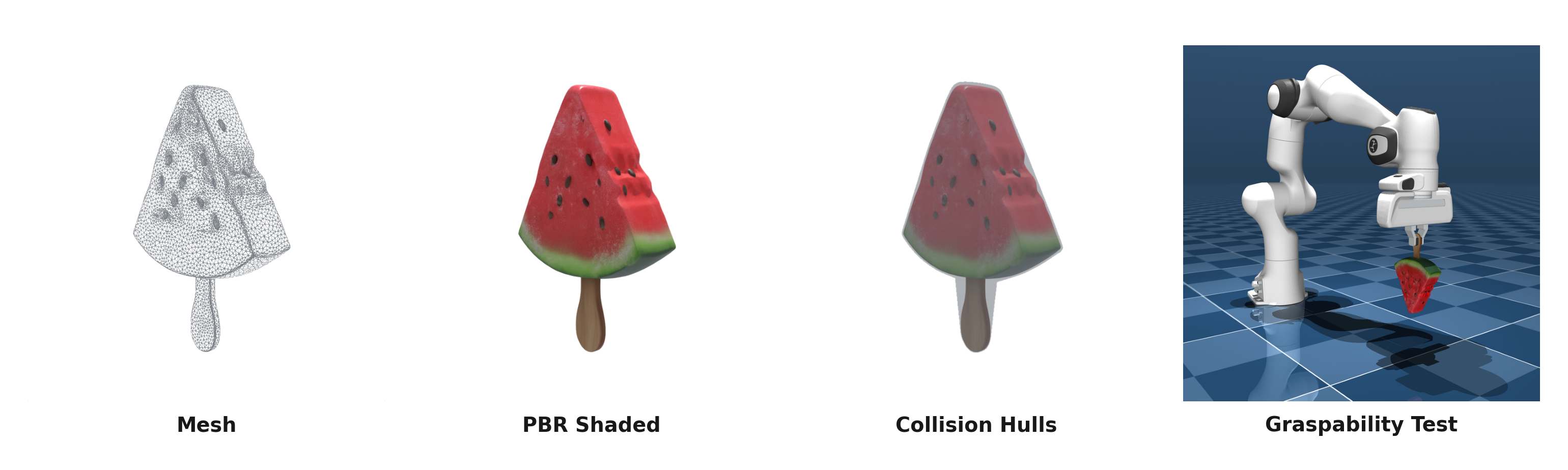}
    \caption[Collision-hull contact conformity]{\textbf{Collision-hull contact conformity (qualitative).} (Left to right): Render mesh, PBR asset, collision hull, and MuJoCo gripper contact. The hull follows the render mesh at the contact points without visible interpenetration. This is not a grasp-success evaluation.}
    \label{fig:grasping_showcase}
\end{figure}

Task-level planners often attach objects to the gripper kinematically, bypassing collision meshes, whereas physics-based contact depends on the asset's hull. Figure~\ref{fig:grasping_showcase} shows shipped AmaraSpatial-10K collision hulls in contact with a simulated gripper in MuJoCo. The figure illustrates that the hull follows the render mesh closely at the contact points. It does not measure grasp success, which depends jointly on the planner and the object geometry; in preliminary trials we could not separate planner failures from asset effects, so we report no success rate (\S\ref{sec:discussion}).

\FloatBarrier

\section{Dataset Generation Pipeline}
\label{app:generation_pipeline}

\subsection{Taxonomy and Description Generation}

A dual-LLM pipeline (Qwen-32B~\cite{qwen} and Gemini 3 via
API~\cite{gemini}\footnote{Gemini 3 accessed via the public API in
March--April 2026; Qwen2-32B-Instruct checkpoint dated 2024-09.}) defines asset
specifications prior to 3D synthesis. For each of the 476 subcategories, the
pipeline produces a multi-sentence description (style, material, era,
functional context), a reference-image prompt, and real-world dimension
estimates in metres. The process is seeded per subcategory for reproducibility.
2D reference images are then generated from the prompts using Gemini 3 Flash
Image~\cite{gemini,nanobanana}.

\subsection{3D Mesh Generation and Standardisation}
Core geometries are produced by an in-house generation
engine\footnote{\url{https://amara.01c.ai}} conditioned on the LLM text and
reference image. Architectural details are out of scope; we report measured
asset properties (\S\ref{sec:intrinsic}) rather than defend the underlying
method. Raw meshes undergo automatic retopology and standardisation along
three axes: \textbf{polycount} (most assets target $\sim$50{,}000 triangles,
with $\sim$2{,}000 low-poly assets at $\sim$10{,}000 and $\sim$1{,}000
``hero'' assets at $\sim$100{,}000), \textbf{PBR materials} (high-frequency
geometric detail baked into separated Normal and Roughness maps embedded in
the \texttt{.glb}, preserving relighting fidelity), and \textbf{collision
hulls} (a convex hull below 1{,}000 triangles per asset, serving as a
real-time physics proxy).

\subsection{Spatial Alignment}
\label{app:spatial_alignment}

Each raw mesh $V_{\text{raw}} \in \mathbb{R}^{N \times 3}$ is transformed via
$V_{\text{aligned}} = s \cdot (R \cdot V_{\text{raw}}) + \mathbf{t}$, where:

\begin{itemize}
    \item \textbf{Metric scaling ($s$).} The scale factor matches the asset's
    primary dimension to the LLM-estimated real-world size (e.g.\ a dining
    chair with estimated height 0.85\,m yields $s$ such that
    $\max(V_z) - \min(V_z) = 0.85$ after rotation). Scale accuracy is
    validated via SPS (\S\ref{sec:scale_plausibility}).
    \item \textbf{Axis-aligned rotation ($R$).} An off-the-shelf VLM renders
    each asset from the $+X$ direction and classifies whether the functional
    front is visible. Assets classified as front-facing are accepted; others
    are rotated in $90^\circ$ increments until a front-facing render is
    confirmed, or flagged for manual inspection.
    \item \textbf{Semantic anchoring ($\mathbf{t}$).} The origin is placed at
    bottom-centre ($Z_{\min}$) for ground-resting objects, top-centre
    ($Z_{\max}$) for ceiling-mounted objects, and volumetric centroid for
    suspended objects.
\end{itemize}

\subsection{Curation Protocol}
\label{app:curation}

Each candidate asset passes three automated gates before inclusion: (i)
\emph{geometric health} (manifoldness, non-degenerate triangle fraction,
polycount within band); (ii) \emph{scale plausibility} (measured primary
dimension within a generous $[\ell/3,\,3u]$ envelope of the LLM-judged
interval, rejecting catastrophic mis-scaling while retaining stylistic
variation); and (iii) \emph{forward-axis audit} (VLM classification
confirming front-facing orientation). Assets failing any gate are auto-repaired
where possible or discarded.

Overall, 91.4\% of candidates pass all three gates. Of the 8.6\% rejected:
3.1\% fail the geometric-health gate, 4.0\% the scale-plausibility gate, and
1.5\% the forward-axis audit (assets failing multiple gates are counted at the
first failure). No deduplication is applied beyond asset-ID uniqueness;
near-duplicates within a subcategory are retained as legitimate stylistic
variation. A geometry-based audit finds that they are rare: 11 of the 10{,}049 assets in multi-asset subcategories (0.11\%) are near-duplicates, and asset IDs allow users to deduplicate if they wish. AmaraSpatial-10K is released as an asset gallery rather than a
supervised benchmark, so no canonical train/val/test split is provided;
consumers training on the dataset should hold out a subset stratified by
subcategory.

\section{LLM-as-Judge Prompts}
\label{app:llm_prompts}

To derive plausible dimension intervals $[\ell, u]$ for each semantic category, we use an LLM-as-Judge protocol. We employ two complementary prompting modes (Text and Vision) and aggregate the results via a three-run union strategy to ensure robustness against degenerate responses.

\paragraph{Interval Derivation and Three-Run Union Protocol.}
Rather than asking the LLM to guess an arbitrary range, the model is prompted to provide a single typical maximum dimension $d$ (in centimeters) for a given object. Each query is issued independently three times (temperature $T=0.1$), yielding three point estimates: $d_1, d_2, d_3$. 

Each estimate $d_i$ is converted to meters and expanded into a plausible interval $[\ell_i, u_i] = [0.7 d_i, 1.3 d_i]$, providing a $\pm30\%$ tolerance band around the judged typical size. The final interval for the category is defined as the union of these three runs:
\begin{equation*}
    [\ell, u] = [\min(\ell_1, \ell_2, \ell_3),\; \max(u_1, u_2, u_3)]
\end{equation*}
This union strategy significantly reduces the risk of a single overly specific response improperly narrowing the valid scale range. For text-mode queries (used for named categories), the prompt in Box~\ref{box:scale_text} is utilized. For categories where visual examples are available, the vision-mode prompt in Box~\ref{box:scale_vision} is used instead, with Gemini 2.5 Flash serving as the judge model.

\vspace{1em}

\refstepcounter{promptbox}\label{box:scale_text}
\begin{tcolorbox}[
    title={\textbf{Box \thepromptbox: Scale Validation Prompt Template (Text Mode)}},
    colback=gray!5, colframe=black!70,
    fonttitle=\bfseries\small, fontupper=\ttfamily\small,
    breakable
]
You are a precise database of physical object length dimensions.\\
Respond ONLY with the maximum dimension (length, width, or height) of the\\
given real-world object in centimeters.\\
Provide a realistic, common value as an integer. Do not write any other words.\\[6pt]
Object: soda can\\
Target Longest Dimension (centimeters): 12\\[4pt]
Object: car\\
Target Longest Dimension (centimeters): 450\\[4pt]
Object: office chair\\
Target Longest Dimension (centimeters): 110\\[4pt]
Object: standard king bed\\
Target Longest Dimension (centimeters): 200\\[4pt]
Object: table\\
Target Longest Dimension (centimeters): 150\\[4pt]
Object: shotgun\\
Target Longest Dimension (centimeters): 100\\[4pt]
Object: \textit{\{category\_name\}}\\
Target Longest Dimension (centimeters):
\end{tcolorbox}

\refstepcounter{promptbox}\label{box:scale_vision}
\begin{tcolorbox}[
    title={\textbf{Box \thepromptbox: Scale Validation Prompt Template (Vision Mode)}},
    colback=gray!5, colframe=black!70,
    fonttitle=\bfseries\small, fontupper=\ttfamily\small,
    breakable
]
Analyze this image and provide:\\
1.\ A brief description of the object shown (1--2 sentences).\\
2.\ The realistic maximum dimension (length, width, or height --- whichever\\
\phantom{2.\ }is largest) this object would have in the real world, in centimeters.\\[6pt]
Respond in this exact format:\\
DESCRIPTION: [your description here]\\
MAX\_SIZE\_CM: [integer number only]\\[6pt]
Examples of realistic sizes:\\
- A soda can: 12\,cm\\
- A car: 450\,cm\\
- An office chair: 110\,cm\\
- A dining table: 150\,cm\\
- A smartphone: 16\,cm\\
- A house: 1000\,cm\\
- A person: 180\,cm\\[6pt]
Be precise and realistic with the size estimation based on what the object actually is.
\end{tcolorbox}

\section{SPS Sensitivity Analysis}
\label{app:sps_sensitivity}

To verify that the Scale Plausibility Score rankings are not artefacts of the
Gaussian decay shape, we repeat the evaluation with two alternative decay functions.
Let $d = \max(0,\, \ell - x) + \max(0,\, x - u)$ be the signed distance of height
$x$ outside interval $[\ell, u]$, and $h = (u - \ell)/2$ the half-width.
The three functions are:

\begin{align}
  \text{Gaussian (Eq.~\ref{eq:sps}):} \quad  & f_G(x) = \exp\!\left(-\tfrac{d^2}{h^2}\right) \\
  \text{Linear:}   \quad & f_L(x) = \max\!\left(0,\; 1 - \tfrac{d}{h}\right) \\
  \text{Lorentzian:}\quad & f_{\mathcal{L}}(x) = \frac{1}{1 + (d/h)^{2}}
\end{align}

All three functions equal 1.0 when $x \in [\ell, u]$ (i.e.\ \% Perfect is invariant
to decay choice). They differ only in how sharply they penalise out-of-interval assets.

\begin{table}[ht]
\centering
\small %
\setlength{\tabcolsep}{4pt} %
\aboverulesep=0pt
\belowrulesep=0pt
\caption{SPS under alternative decay functions for AmaraSpatial-10K.
         Columns show mean SPS per category; \textbf{Rank} columns show the
         ordinal ranking (1 = highest). Rankings are fully consistent across
         all three decay functions, confirming that the choice of $f$ does not
         affect relative ordering.}
\label{tab:sps_sensitivity}
\resizebox{\linewidth}{!}{%
\begin{tabular}{l c r >{\columncolor{heavenlygold}}c c c r r r}
\toprule
\textbf{Category} &
\textbf{$[\ell, u]$ (m)} &
\textbf{$N$} &
\textbf{\makecell{Gaussian\\$f_G$}} &
\textbf{\makecell{Linear\\$f_L$}} &
\textbf{\makecell{Lorentzian\\$f_{\mathcal{L}}$}} &
\textbf{\makecell{Rank\\$f_G$}} &
\textbf{\makecell{Rank\\$f_L$}} &
\textbf{\makecell{Rank\\$f_{\mathcal{L}}$}} \\
\midrule
Architecture      & 3.0--100.0  &  733 & 0.988 & 0.900 & 0.950 & 1 & 1 & 1 \\
Nature (Flora)    & 0.1--20.0   &  638 & 0.981 & 0.890 & 0.940 & 2 & 2 & 2 \\
Storage Furniture & 0.5--2.4    &  300 & 0.980 & 0.880 & 0.930 & 3 & 3 & 3 \\
Animal            & 0.2--3.0    &  743 & 0.904 & 0.800 & 0.880 & 4 & 4 & 4 \\
Seating           & 0.6--1.1    &  353 & 0.812 & 0.720 & 0.790 & 5 & 5 & 5 \\
Electronics       & 0.05--0.9   &  207 & 0.768 & 0.680 & 0.750 & 6 & 6 & 6 \\
Vehicle           & 1.0--3.5    & 1101 & 0.762 & 0.670 & 0.740 & 7 & 7 & 7 \\
Table / Desk      & 0.4--0.9    &  558 & 0.672 & 0.580 & 0.650 & 8 & 8 & 8 \\
Tableware         & 0.05--0.30  &  589 & 0.479 & 0.400 & 0.450 & 9 & 9 & 9 \\
\midrule
\textbf{Overall}  & ---         & \textbf{5{,}222} &
  \textbf{0.815} & \textbf{0.728} & \textbf{0.787} &
  \multicolumn{3}{c}{\textit{Kendall's $\tau = 1.00$}} \\
\bottomrule
\end{tabular}}
\end{table}
\FloatBarrier

As Table~\ref{tab:sps_sensitivity} shows, the rank order is identical across all three decay functions, confirming that all qualitative conclusions in the paper are robust to the choice of decay function.

\section{Asset Usability in Procedural Scene Composition}
\label{app:scene_composition}

\begin{figure}[H]
\centering
\includegraphics[width=0.85\textwidth]{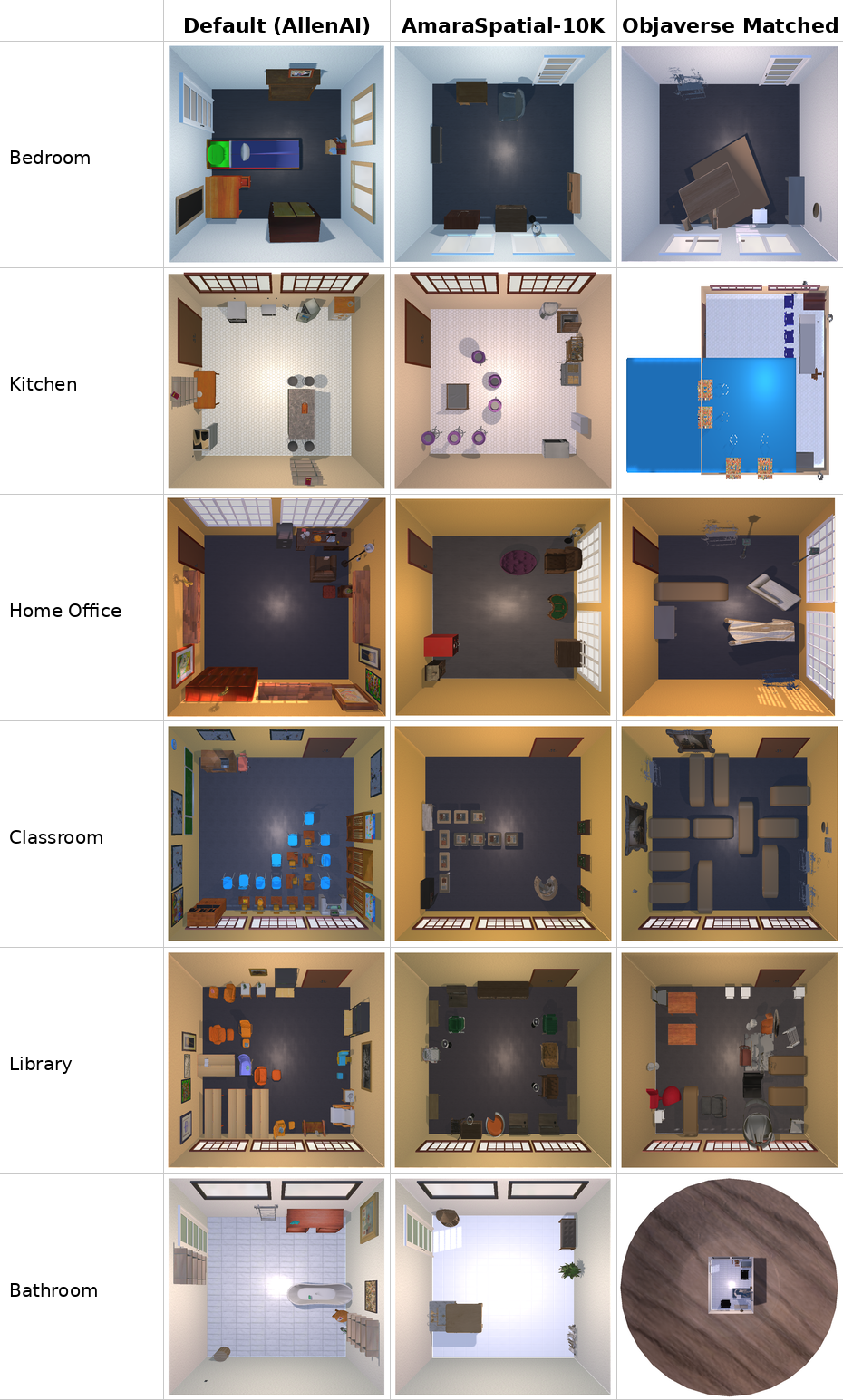}
\caption{Top-down renders of 6 scenes across all three asset packs from the original six-scene study. Each row is a scene generated from the same prompt, with each column using a different asset pack. AmaraSpatial-10K and Default renders are visually comparable, while the Objaverse Matched pack shows more variation in object placement and rendering quality.}
\label{fig:scene_panel}
\end{figure}

To enable reproduction of the Holodeck evaluation reported in \S\ref{sec:asset-usabilty} and to permit visual inspection of the generated scenes, Table~\ref{tab:prompts} provides the original six scene-composition prompts and Figure~\ref{fig:scene_panel} shows top-down renders of the resulting scenes across all three asset packs. Table~\ref{tab:quality_results_original} reports the original six-scene, three-pack study from our submission, including Holodeck's Default (AllenAI) pack, which uses Allen AI's \texttt{objathor} preprocessing, and Gemini~2.5~Flash perceptual scores. Its Objaverse Matched column is affected by the conversion error described in \S\ref{sec:asset-usabilty}, so its geometric gaps should not be read as source effects; the corrected comparison is Table~\ref{tab:quality_results}.

\begin{table}[ht]
\centering
\small
\setlength{\tabcolsep}{3.5pt}
\renewcommand{\arraystretch}{1.06}
\aboverulesep=0pt
\belowrulesep=0pt
\caption{\textbf{Original study (superseded): scene quality across asset packs.} Values are mean $\pm$ standard deviation ($N{=}6$ scenes per pack). Geometry: scene JSON; perceptual: Gemini~2.5~Flash ratings (1--10). The Objaverse Matched column predates the conversion fix; see Table~\ref{tab:quality_results} for the corrected comparison.}
\label{tab:quality_results_original}
\begin{tabular}{@{}c l c >{\columncolor{heavenlygold}}c c@{}}
\toprule
& \textbf{Metric} & \textbf{\makecell{Default\\(AllenAI)}} & \textbf{AmaraSpatial-10K} & \textbf{\makecell{Objaverse\\Matched}} \\
\midrule
\multirow{5}{*}{\rotatebox[origin=c]{90}{\footnotesize\textit{Geometric}}}
& Object Overlap $\downarrow$     & $2.00 \pm 1.26$  & $\mathbf{0.00 \pm 0.00}$   & $3.83 \pm 3.06$ \\
& Containment (\%) $\uparrow$     & $100.00 \pm 0.00$ & $100.00 \pm 0.00$ & $100.00 \pm 0.00$ \\
& Floor Contact (\%) $\uparrow$   & $71.83 \pm 15.83$ & $\mathbf{100.00 \pm 0.00}$ & $68.88 \pm 18.18$ \\
& Scale Consistency $\uparrow$    & $0.69 \pm 0.09$  & $0.71 \pm 0.10$   & $\mathbf{0.76 \pm 0.14}$ \\
& Spacing Regularity $\uparrow$   & $0.58 \pm 0.17$  & $0.58 \pm 0.02$   & $\mathbf{0.66 \pm 0.13}$ \\
\midrule
\multirow{3}{*}{\rotatebox[origin=c]{90}{\footnotesize\textit{Percept.}}}
& Facing Relationships $\uparrow$ & $\mathbf{7.17 \pm 1.72}$  & $7.00 \pm 2.00$   & $6.00 \pm 3.22$ \\
& Grouping Coherence $\uparrow$   & $\mathbf{6.50 \pm 1.52}$  & $5.67 \pm 2.34$   & $5.33 \pm 3.14$ \\
& Compositional Harmony $\uparrow$& $4.83 \pm 1.17$  & $\mathbf{5.00 \pm 2.53}$   & $4.17 \pm 2.23$ \\
\bottomrule
\end{tabular}
\end{table}

\begin{table}[ht]
\centering
\caption{Evaluation prompts for scene generation in the original six-scene study. Each prompt specifies a room type with key furniture items and spatial relationships.}
\label{tab:prompts}

\begin{tabular}{@{}clp{8cm}@{}}
\toprule
\# & Scene Type & Prompt \\
\midrule
1 & Bedroom & A small bedroom with a wooden desk against the wall and a single bed with a nightstand \\
2 & Kitchen & A modern kitchen with a central island, bar stools, and stainless steel appliances along the back wall \\
3 & Home Office & A home office with floor-to-ceiling bookshelves, a large desk facing the window, and a leather armchair \\
4 & Classroom & A classroom with rows of student desks facing a whiteboard, a teacher desk in the front corner, and a globe on a shelf \\
5 & Library & A library reading room with long wooden tables, desk lamps, and tall bookshelves lining every wall \\
6 & Bathroom & A bathroom with a freestanding bathtub, a vanity with double sinks, and a towel rack beside the shower \\
\bottomrule
\end{tabular}
\end{table}
\FloatBarrier

\begin{table}[ht]
\centering
\small
\renewcommand{\arraystretch}{1.04}
\caption{Scene quality metrics. Geometric metrics are computed deterministically from the scene JSON. Perceptual metrics are scored 1--10 by Gemini 2.5 Flash from the top-down render.}
\label{tab:metrics_definitions}
\begin{tabular}{@{}lp{7.8cm}@{}}
\toprule
\textbf{Metric} & \textbf{Description} \\
\midrule
\multicolumn{2}{@{}l}{\textit{Geometric}} \\
Object Overlap $\downarrow$ & Number of floor object pairs with intersecting 2D bounding polygons ($>1\text{cm}^2$ threshold). \\
Containment $\uparrow$ & Percentage of objects whose XZ position falls within their assigned room's floor polygon. \\
Floor Contact $\uparrow$ & Percentage of floor objects with Y-position in the plausible range $[0, 1.5]$m. \\
Scale Consistency $\uparrow$ & Mean SPS comparing planned object dimensions against category-specific real-world size ranges (e.g.\ chairs 0.7--1.1m). Range 0--1. \\
Selected-vs-planned scale $\uparrow$ & Agreement between the dimensions of each retrieved asset and the dimensions in Holodeck's layout plan. Range 0--1. \\
Floor-anchored at ingest $\uparrow$ & Percentage of assets whose origin already lies at the floor contact point when loaded, before any re-anchoring by the pipeline. \\
Spacing Regularity $\uparrow$ & For groups of $\geq$3 identical assets, inverted coefficient of variation of pairwise distances. Range 0--1. \\
\midrule
\multicolumn{2}{@{}l}{\textit{Perceptual}} \\
Facing Relationships $\uparrow$ & Whether seating objects face tables or each other; desks face walls or windows. \\
Grouping Coherence $\uparrow$ & Whether functionally related objects are placed near each other (e.g.\ nightstand beside bed). \\
Compositional Harmony $\uparrow$ & Overall balance, realism, and aesthetic plausibility of the layout. \\
\bottomrule
\end{tabular}
\end{table}

\subsection{Asset Pack Pipeline Details}
\label{app:pipeline}

This appendix documents the full processing pipeline behind the three asset packs compared in \S\ref{sec:asset-usabilty}.

\paragraph{Default (AllenAI).} 50{,}092 Objaverse 1.0 objects preprocessed via Allen AI's \texttt{objathor} pipeline (Blender-based transform baking, geometry normalization, texture extraction) into THOR-native \texttt{.pkl.gz} format, plus 1{,}371 THOR base objects covering doors, windows, and other structural elements. Each asset ships with CLIP ViT-L-14 image embeddings (3 views, 768-d), SBERT \texttt{all-mpnet-base-v2} text embeddings (768-d), and LLM-generated category labels, descriptions, bounding boxes, and placement metadata.

\paragraph{AmaraSpatial-10K.} For Holodeck integration, we computed CLIP image embeddings (ViT-L-14, front/left/right renders), SBERT text embeddings (\texttt{all-mpnet-base-v2}) from asset descriptions, and converted GLBs to THOR format using the same converter as the Objaverse Matched pack. We used an empty THOR base directory so scenes draw furniture and decor exclusively from AmaraSpatial-10K, with only structural elements supplied from the THOR base set.

\paragraph{Objaverse Matched.} A curated indoor subset of Objaverse 1.0 built in five stages: (i)~keyword matching against 80+ indoor terms across 8 semantic groups (seating, tables, storage, beds, lighting, decor, kitchenware, bathroom), yielding ${\sim}215\text{K}$ candidates; (ii)~reproducible random sampling to 22K assets (seed = 42); (iii)~scale-plausibility filtering using
\begin{equation}
\text{SPS} = \exp\!\left(-\left(d/h\right)^2\right),
\end{equation}
where $d$ is the distance to the nearest category-specific size-range boundary and $h$ its half-width, rejecting $\text{SPS} < 0.3$ to yield 10{,}417 assets; (iv)~GLB-to-THOR conversion (mesh, PBR textures, bounding-box colliders, visibility points); (v)~CLIP (ViT-L-14) and SBERT (\texttt{all-mpnet-base-v2}) embeddings from descriptions.

\paragraph{Residual differences from AmaraSpatial-10K.} Two minor differences remain between the AmaraSpatial-10K and Objaverse Matched packs: text-derived rather than image-derived CLIP embeddings, and bounding-box rather than dedicated collision meshes. Neither is expected to materially affect results, since Holodeck blends CLIP and SBERT scores during retrieval and uses bounding-box colliders for placement throughout.

\newpage

\FloatBarrier

\newpage
\section*{NeurIPS Paper Checklist}

\begin{enumerate}

\item {\bf Claims}
    \item[] Question: Do the main claims made in the abstract and introduction accurately reflect the paper's contributions and scope?
    \item[] Answer: \answerYes{}
    \item[] Justification: The abstract and Section~\ref{sec:intro} state the dataset's properties (metric scale, semantic anchoring, PBR materials, collision hulls, paired reference image, multi-sentence text), enumerate three contributions, and quote the headline results ($3.4\times$ CLIP R@5, $99.1\%$ settling stability, better planned-scale matching and a blinded human preference in Holodeck scenes); each is substantiated in Sections~\ref{sec:intrinsic}--\ref{sec:relevance-robotics}. The scope is stated explicitly: the evidence supports standardised ingestion, static placement, and basic physics compatibility, and manipulation and model training are not evaluated.

\item {\bf Limitations}
    \item[] Question: Does the paper discuss the limitations of the work performed by the authors?
    \item[] Answer: \answerYes{}
    \item[] Justification: Section~\ref{sec:discussion} (Limitations) addresses the synthetic-only nature of the assets, non-uniform category coverage with 23 single-asset subcategories, the restricted PBR encoding (Normal/Roughness only), the resolution limits of category-level SPS intervals, known CLIP compositional failure modes, English-only metadata, and the fact that metric dimensions are LLM-estimated rather than physically measured (calibrated only for household categories), the narrow scope of the physics evaluation (no manipulation), the absence of an image-to-3D training experiment and of an end-to-end ingestion-time study, and the modest size of the scene-composition study.

\item {\bf Theory assumptions and proofs}
    \item[] Question: For each theoretical result, does the paper provide the full set of assumptions and a complete (and correct) proof?
    \item[] Answer: \answerNA{}
    \item[] Justification: The paper introduces a dataset and an empirical evaluation suite; it contains no formal theorems or proofs. The single defined quantity (SPS, Eq.~\ref{eq:sps}) is a metric definition rather than a theoretical result, and its sensitivity to the decay choice is reported empirically in Appendix~\ref{app:sps_sensitivity}.

    \item {\bf Experimental result reproducibility}
    \item[] Question: Does the paper fully disclose all the information needed to reproduce the main experimental results of the paper to the extent that it affects the main claims and/or conclusions of the paper (regardless of whether the code and data are provided or not)?
    \item[] Answer: \answerYes{}
    \item[] Justification: The intrinsic evaluation suite is fully specified in Section~\ref{sec:intrinsic} (SPS formula in Eq.~\ref{eq:sps}, LLM-as-Judge protocol in \S\ref{sec:sps_validation}, CLIP ViT-L/14 four-view rendering protocol in \S\ref{sec:clip_coherence}, Concept Density keyword banks in \S\ref{sec:semantic_richness}). Downstream protocols are described for retrieval (\S\ref{sec:benchmark_retrieval}), Holodeck scene composition (\S\ref{sec:asset-usabilty} and Table~\ref{tab:prompts}), and Habitat-Sim physics (\S\ref{sec:relevance-robotics}). The dataset generation and curation pipelines are described in Appendices~\ref{app:generation_pipeline} and~\ref{app:curation}, the prompt templates in Appendix~\ref{app:llm_prompts}, and the dataset is publicly released on Hugging Face under CC~BY~4.0 with reference implementations of the evaluation suite.

\item {\bf Open access to data and code}
    \item[] Question: Does the paper provide open access to the data and code, with sufficient instructions to faithfully reproduce the main experimental results, as described in supplemental material?
    \item[] Answer: \answerYes{}
    \item[] Justification: As stated in Section~\ref{sec:dataset}, AmaraSpatial-10K is publicly available on Hugging Face under CC~BY~4.0 with a five-year maintenance commitment, and reference implementations of the evaluation suite (SPS, anchor error, Concept Density, CLIP coherence) are released alongside the dataset.

\item {\bf Experimental setting/details}
    \item[] Question: Does the paper specify all the training and test details (e.g., data splits, hyperparameters, how they were chosen, type of optimizer) necessary to understand the results?
    \item[] Answer: \answerYes{}
    \item[] Justification: Each evaluation specifies its key settings: the LLM-as-Judge protocol uses three independent queries at $T{=}0.1$ (\S\ref{sec:sps_validation}); CLIP retrieval uses ViT-L/14 with mean-pooled embeddings over four orthographic renders (\S\ref{sec:benchmark_retrieval}); the Holodeck study fixes prompts (Table~\ref{tab:prompts}), two seeds per prompt, CLIP/SBERT encoders, and the curation thresholds for the Objaverse-Matched pack (\S\ref{sec:asset-usabilty}); and the Habitat-Sim drop test specifies the 600-step / 60\,Hz simulation, 10\,cm spawn offset, and pass-criterion thresholds (\S\ref{sec:relevance-robotics}). Section~\ref{sec:dataset} explicitly notes that no canonical train/val/test split is provided and recommends subcategory-stratified hold-outs for downstream consumers.

\item {\bf Experiment statistical significance}
    \item[] Question: Does the paper report error bars suitably and correctly defined or other appropriate information about the statistical significance of the experiments?
    \item[] Answer: \answerYes{}
    \item[] Justification: Standard deviations across the asset population are reported as mean~$\pm$~std for cross-modal CLIP coherence (Table~\ref{tab:clip_coherence}) and Habitat-Sim continuous physics metrics (Table~\ref{tab:robotics_metrics_combined}). The Holodeck study reports paired differences with 95\% confidence intervals over 40 paired scenes, an exact binomial test for the human preference, and Fleiss' $\kappa$ for inter-rater agreement (Table~\ref{tab:quality_results}); the human audit reports observed inter-rater agreement (Table~\ref{tab:external_validation}).

\item {\bf Experiments compute resources}
    \item[] Question: For each experiment, does the paper provide sufficient information on the computer resources (type of compute workers, memory, time of execution) needed to reproduce the experiments?
    \item[] Answer: \answerNo{}
    \item[] Justification: Per-asset wall-clock time for the Habitat-Sim drop test is reported in Table~\ref{tab:robotics_metrics_combined} (\S\ref{sec:relevance-robotics}), but hardware specifications (CPU/GPU type, memory) for the evaluation suite, and total compute for dataset generation, are not disclosed. As acknowledged in the Broader Impact discussion (\S\ref{sec:discussion}), the exact generation-compute figure is withheld because pipeline details are proprietary.

\item {\bf Code of ethics}
    \item[] Question: Does the research conducted in the paper conform, in every respect, with the NeurIPS Code of Ethics \url{https://neurips.cc/public/EthicsGuidelines}?
    \item[] Answer: \answerYes{}
    \item[] Justification: The work releases a synthetic 3D asset dataset of everyday objects, vehicles, architecture, and creatures, with no personal data and no identifiable individuals; the only human involvement is three raters judging assets and rendered scenes; the authors have reviewed the NeurIPS Code of Ethics and confirm conformance.

\item {\bf Broader impacts}
    \item[] Question: Does the paper discuss both potential positive societal impacts and negative societal impacts of the work performed?
    \item[] Answer: \answerYes{}
    \item[] Justification: Section~\ref{sec:discussion} (Broader Impact) discusses the positive impact of lowering the barrier to production-quality 3D asset banks and removing per-asset scale and anchor fixes, and identifies three risks: stylistic concentration in synthetic asset banks, the non-trivial compute footprint of generation, and the need for users to consult per-asset license metadata for commercial deployment.

\item {\bf Safeguards}
    \item[] Question: Does the paper describe safeguards that have been put in place for responsible release of data or models that have a high risk for misuse (e.g., pre-trained language models, image generators, or scraped datasets)?
    \item[] Answer: \answerNA{}
    \item[] Justification: The released asset bank consists of synthetic 3D meshes of furniture, vehicles, architecture, and creatures with associated PBR textures and reference renders; it contains no scraped imagery, no personal data, and no identifiable individuals, and therefore poses no notable misuse risk requiring gated release or safety filters.

\item {\bf Licenses for existing assets}
    \item[] Question: Are the creators or original owners of assets (e.g., code, data, models), used in the paper, properly credited and are the license and terms of use explicitly mentioned and properly respected?
    \item[] Answer: \answerYes{}
    \item[] Justification: The compared datasets (Objaverse, HSSD, ABO, GSO, and TRELLIS-500K, which is discussed) and software dependencies (Holodeck, Habitat-Sim, AI2-THOR/ProcTHOR, CLIP, SBERT, GPT-4o, Gemini, Qwen) are cited at their original venues. AmaraSpatial-10K is released under CC~BY~4.0, as stated in Section~\ref{sec:dataset}.

\item {\bf New assets}
    \item[] Question: Are new assets introduced in the paper well documented and is the documentation provided alongside the assets?
    \item[] Answer: \answerYes{}
    \item[] Justification: AmaraSpatial-10K is documented throughout: Section~\ref{sec:dataset} specifies file formats, metadata fields, license, intended uses, and out-of-scope deployments; Appendix~\ref{app:taxonomy} lists the full subcategory taxonomy with per-subcategory counts; Appendix~\ref{app:generation_pipeline} describes the generation, spatial-alignment, and curation pipelines. Per-asset metadata (taxonomy, metric dimensions, anchor type, forward axis, multi-sentence description) ships with the dataset on Hugging Face.

\item {\bf Crowdsourcing and research with human subjects}
    \item[] Question: For crowdsourcing experiments and research with human subjects, does the paper include the full text of instructions given to participants and screenshots, if applicable, as well as details about compensation (if any)?
    \item[] Answer: \answerYes{}
    \item[] Justification: Three Zero One Creative employees who are not authors of the paper audited 314 assets and rated 40 blinded Holodeck scene pairs (\S\ref{sec:external_validation}, \S\ref{sec:asset-usabilty}). They judged synthetic 3D assets and rendered scenes only, did the work as part of their paid employment, and no personal data were collected.

\item {\bf Institutional review board (IRB) approvals or equivalent for research with human subjects}
    \item[] Question: Does the paper describe potential risks incurred by study participants, whether such risks were disclosed to the subjects, and whether Institutional Review Board (IRB) approvals (or an equivalent approval/review based on the requirements of your country or institution) were obtained?
    \item[] Answer: \answerNA{}
    \item[] Justification: The only human involvement is three company employees rating synthetic 3D assets and rendered scenes as part of their work; no personal data were collected and the task carried no risk to participants, so it is not human-subjects research requiring IRB review.

\item {\bf Declaration of LLM usage}
    \item[] Question: Does the paper describe the usage of LLMs if it is an important, original, or non-standard component of the core methods in this research? Note that if the LLM is used only for writing, editing, or formatting purposes and does \emph{not} impact the core methodology, scientific rigor, or originality of the research, declaration is not required.
    \item[] Answer: \answerYes{}
    \item[] Justification: LLM/VLM usage is declared throughout: Qwen-32B and Gemini~3 are used to author asset metadata, dimensions, and reference-image prompts (Appendix~\ref{app:generation_pipeline}); a separate LLM-as-Judge instance generates plausible scale intervals for SPS (\S\ref{sec:sps_validation}, Appendix~\ref{app:llm_prompts}); a VLM verifies forward-axis orientation during curation (Appendix~\ref{app:spatial_alignment}); GPT-4o drives Holodeck's layout planning and Gemini~2.5 Flash scores perceptual scene quality in the original six-scene study (\S\ref{sec:asset-usabilty}, Appendix~\ref{app:scene_composition}); the LLM/VLM-dependent annotations are checked against measured dimensions and human raters (\S\ref{sec:external_validation}); and CLIP ViT-L/14 is used as a fixed encoder throughout the intrinsic and retrieval evaluations.

\end{enumerate}


\begin{thebibliography}{99}

\bibitem{shapenet} Chang, A.X., et al. ``ShapeNet: An Information-Rich 3D Model Repository.'' \textit{arXiv preprint arXiv:1512.03012}, 2015.

\bibitem{objaverse} Deitke, M., et al. ``Objaverse: A Universe of Annotated 3D Objects.'' \textit{CVPR}, 2023.

\bibitem{objaversexl} Deitke, M., et al. ``Objaverse-XL: A Universe of 10M+ 3D Objects.'' \textit{NeurIPS}, 2023.

\bibitem{gso} Downs, L., et al. ``Google Scanned Objects: A High-Quality Dataset of 3D Scanned Household Items.'' \textit{ICRA}, 2022.

\bibitem{abo} Collins, J., et al. ``ABO: Dataset and Benchmarks for Real-World 3D Object Understanding.'' \textit{CVPR}, 2022.

\bibitem{hssd} Khanna, M., et al. ``Habitat Synthetic Scenes Dataset (HSSD-200): An Analysis of 3D Scene Scale and Realism Tradeoffs for ObjectGoal Navigation.'' \textit{CVPR}, 2024.

\bibitem{habitat} Savva, M., et al. ``Habitat: A Platform for Embodied AI Research.'' \textit{ICCV}, 2019.

\bibitem{igibson} Shen, B., et al. ``iGibson 1.0: A Simulation Environment for Interactive Tasks in Large Realistic Scenes.'' \textit{IROS}, 2021.

\bibitem{procthor} Deitke, M., et al. ``ProcTHOR: Large-Scale Embodied AI Using Procedural Generation.'' \textit{NeurIPS}, 2022.

\bibitem{hm3d} Ramakrishnan, S.K., et al. ``Habitat-Matterport 3D Dataset (HM3D): 1000 Large-scale 3D Environments for Embodied AI.'' \textit{NeurIPS Datasets and Benchmarks}, 2021.

\bibitem{matterport3d} Chang, A., et al. ``Matterport3D: Learning from RGB-D Data in Indoor Environments.'' \textit{3DV}, 2017.

\bibitem{triposr} Tochilkin, D., et al. ``TripoSR: Fast 3D Object Reconstruction from a Single Image.'' \textit{arXiv preprint arXiv:2403.02151}, 2024.

\bibitem{instantmesh} Xu, J., et al. ``InstantMesh: Efficient 3D Mesh Generation from a Single Image with Sparse-view Large Reconstruction Models.'' \textit{arXiv preprint arXiv:2404.07191}, 2024.

\bibitem{lrm} Hong, Y., et al. ``LRM: Large Reconstruction Model for Single Image to 3D.'' \textit{ICLR}, 2024.

\bibitem{crm} Wang, Z., et al. ``CRM: Single Image to 3D Textured Mesh with Convolutional Reconstruction Model.'' \textit{ECCV}, 2024.

\bibitem{orientationmatters} Lu, Y., et al. ``Orientation Matters: Making 3D Generative Models Orientation-Aligned.'' \textit{NeurIPS}, 2025.

\bibitem{canoverse} Jin, L., et al. ``CanoVerse: 3D Object Scalable Canonicalization and Dataset for Generation and Pose.'' \textit{arXiv preprint arXiv:2603.07144}, 2026.

\bibitem{trellis} Xiang, J., et al. ``Structured 3D Latents for Scalable and Versatile 3D Generation.'' \textit{CVPR}, 2025.

\bibitem{vhacd} Mamou, K. and Ghorbel, F. ``A Simple and Efficient Approach for 3D Mesh Approximate Convex Decomposition.'' \textit{IEEE International Conference on Image Processing (ICIP)}, 2009.

\bibitem{fleiss} Fleiss, J.L. ``Measuring Nominal Scale Agreement Among Many Raters.'' \textit{Psychological Bulletin}, 76(5):378--382, 1971.

\bibitem{holodeck} Yang, Y., et al. ``Holodeck: Language Guided Generation of 3D Embodied AI Environments.'' \textit{CVPR}, 2024.

\bibitem{layoutgpt} Feng, W., et al. ``LayoutGPT: Compositional Visual Planning and Generation with Large Language Models.'' \textit{NeurIPS}, 2023.

\bibitem{clip} Radford, A., et al. ``Learning Transferable Visual Models From Natural Language Supervision.'' \textit{ICML}, 2021.

\bibitem{gpt4o}
OpenAI. GPT-4o System Card. \textit{arXiv preprint arXiv:2410.21276}, 2024.

\bibitem{sbert}
Reimers, N. and Gurevych, I. Sentence-BERT: Sentence Embeddings using Siamese BERT-Networks. In \textit{Proceedings of the 2019 Conference on Empirical Methods in Natural Language Processing (EMNLP)}, 2019.

\bibitem{qwen}
Yang, A., et al. ``Qwen2 Technical Report.'' \textit{arXiv preprint arXiv:2407.10671}, 2024.

\bibitem{gemini}
Gemini Team, Google. ``Gemini: A Family of Highly Capable Multimodal Models.'' \textit{arXiv preprint arXiv:2312.11805}, 2023.

\bibitem{nanobanana}
Google DeepMind. ``Gemini 3 Flash Image (Nano Banana 2).'' \url{https://blog.google/innovation-and-ai/technology/ai/nano-banana-2/}, 2025. Accessed April 23, 2026.
\end{thebibliography}
\end{document}